\documentclass[preprint,12pt]{myarticle}

\usepackage{amssymb}
\usepackage{xcolor}
\usepackage{amsmath,amsfonts}
\usepackage[caption=false,font=scriptsize,labelfont={scriptsize,bf},textfont=sf]{subfig}
\usepackage{hyperref}
\usepackage{tabularx}
\usepackage{multirow}

\journal{Pattern Recognition}

\begin{document}

\begin{frontmatter}



\title{2D image head pose estimation via latent space regression under occlusion settings}





\author{José Celestino}
\ead{Corresponding author - jose.celestino@tecnico.ulisboa.pt}
\author{Manuel Marques}
\ead{manuel@isr.tecnico.ulisboa.pt}
\author{Jacinto C. Nascimento}
\ead{jan@isr.tecnico.ulisboa.pt}
\author{João Paulo Costeira}
\ead{jpc@isr.tecnico.ulisboa.pt}
\address{Institute for Systems and Robotics, Instituto Superior Técnico, Lisboa, Portugal}




\begin{abstract}
\par
Head orientation is a challenging Computer Vision problem that
has been extensively researched having a wide variety of applications. However, current state-of-the-art systems still underperform in the presence of
occlusions and are unreliable for many task applications in such scenarios. This work proposes a novel deep learning approach for the problem of head pose estimation under occlusions.
The strategy is based on latent space regression as a fundamental key
to better structure the problem for occluded scenarios. Our model surpasses several state-of-the-art methodologies for occluded HPE, and achieves similar accuracy for non-occluded scenarios.  We demonstrate the usefulness of the proposed approach with: (i) two synthetically occluded versions of the BIWI and  AFLW2000 datasets, (ii) real-life occlusions of the Pandora dataset, and (iii) a real-life application to human-robot interaction scenarios where face occlusions often occur. Specifically, the autonomous feeding from a robotic arm.

\end{abstract}



\end{frontmatter}


\section{Introduction}
Extensively researched over the last 25 years \cite{ref1,ref44}, 2D head pose estimation (HPE) is a challenging but compelling and relevant computer vision problem, since several applications benefit from knowing the position of the head. This includes driving safety systems \cite{ref2, ref38}, motion capture \cite{ref3}, human behavior analysis in smart rooms to monitor activities \cite{ref35}, human interaction in meeting and workplaces \cite{ref36, ref37}, surveillance and safety \cite{ref39} and target advertisement \cite{ref40}. Succinctly, this problem consists in approximately determining the orientation of a head in a 2D image. 
\par
\begin{figure}[h] 
    \centering
  \subfloat[Driver attention systems]{%
       \includegraphics[height=0.25\linewidth]{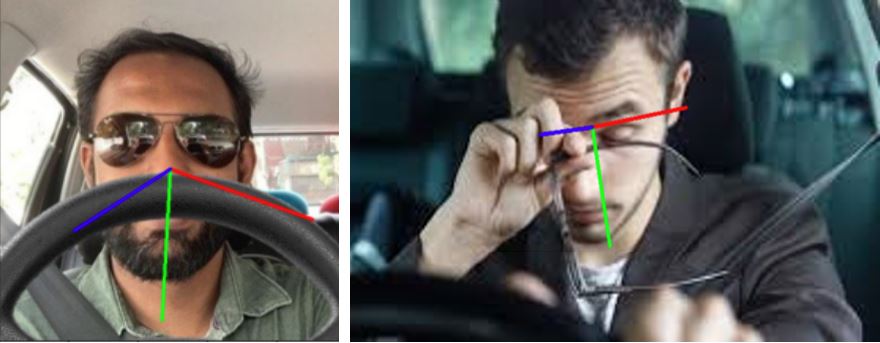}}
    \hfill
  \subfloat[Autonomous assisted feeding]{%
        \includegraphics[height=0.25\linewidth]{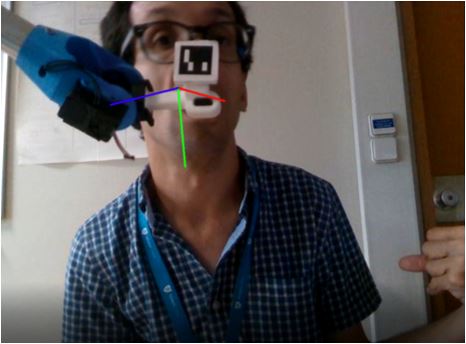}}
  \caption{Head pose problem for occlusion scenarios (blue axis points towards estimated face direction).}
  \label{fig1} 
\end{figure}

 Deep learning has become a fundamental tool to address this problem reaching state-of-the-art (SoTA) performances. However, current deep learning based systems scarcely approach the HPE under the facial occlusions. This is a fundamental challenge that must be somewhat addressed, since occlusions emerge from  numerous applications ({\em e.g.} see Fig.~\ref{fig1}). As it will be addressed herein, a focus is given to scenarios where occlusions are typical and abundant, an issue that is seldom in most recent HPE approaches.

To address the above challenge we present a novel deep learning methodology based on latent space regression that enables to tackle the problem of head occlusions. As per the hierarchical representation of HPE techniques and applications defined in \cite{ref44}, our work is aligned with a CNN-based
training method which has applications for problems that involve 2D images of low or high resolution, 
such as driver attention \cite{ref2}, surveillance \cite{ref39} and face frontalization \cite{ref49}.
\par Our purpose is twofold. Particularly, we aim  (i)  to achieve robust 2D head pose estimation for occluded faces, and (ii)  to extend on current works that achieve SoTA estimation results in non-occluded benchmark datasets.    

\par
The contributions can be highlighted as follows\footnote{Code available at \url{http://github.com/Jose-Celestino/OcclusionHPE}}:


\begin{itemize}
    \item A theoretical learning framework for HPE relying on latent space regression with multi-loss.
    \item An ablation study reporting how the above innovation impacts on the HPE estimates.
    \item Achievement of SoTA accuracy results for HPE in several benchmark datasets with occlusions. As a side result we also demonstrate the effectiveness of the proposed approach which surpasses SoTA in scenarios without occlusion.
    \item A new application for HPE. Specifically, the ability to feed disabled  people using a robotic arm.
    \item A simple auxiliary approach, yet effective, to generate synthetic occlusions. In this way, it is possible to build new (occluded) datasets with any desired occlusion object.
\end{itemize}



\section{Related work}


In this section we describe human head pose estimation related literature regarding model based and learning based approaches. 

\subsection{Model-based strategies}

Model-based methods aim to recover the head motion from one (or more) camera views using a, {\it e.g.}, 2D rigid model. Such a model is usually parameterized with a kinematic chain and thereby the pose is represented by a vector of landmarks ({\it i.e.}, points) and the corresponding joint angles. The goal is to fit the model to observations ({\it i.e.}, features) collected from the image.

The authors from \cite{ref6} propose a method that performs face landmark and keypoint prediction. Landmarks represent a specific set face points ({\it  e.g.}, eyes, nose) that define a rigid model. Keypoints are feature head points extracted and tracked throughout frames. The number of keypoints per frame is not ﬁxed as in the facial landmarks, since it depends on how many are tracked between two frames. They also do not have fixed locations. The authors use the Features from Accelerated Segment Test (FAST) and a pyramidal Lucas-Kanade feature tracker to detect and track these keypoints. They use a Kalman filter to blend both keypoint (prediction step) and landmark detection (correction step). This fusion method shows better results than either using only keypoints or landmarks. 
\par In \cite{ref65} the authors propose a geometry-based method capable of estimating the head pose from a single 2D face image which only requires four non-coplanar 2D feature points. The idea of using less feature points is to  reduce computational complexity . They perform feature point normalization and a 3D morphing method with spherical parametrization to adapt the 3D facial model to each individual and counter generalization issues. 

\par  The idea in \cite{dad3d} is to implement a model that maps an input image to a 3D FLAME head mesh representation and simultaneously learns the head pose, shape, and expression. The training methodology is landmark-based via optimization of two loss components, a reprojection loss and a shape plus expression loss. The first one is based on the full head dense 2D landmarks and aims to assess the pose accuracy, while the second measures how well the predicted 3D head mesh fits. The authors also developed a dataset of nearly 45k images to train and validate the model, but only 16\% of them include occlusions.

\par The work developed in \cite{ref11} approaches the 3D face alignment and pose estimation problems as a 3D Morphable Model (3DMM) parameter regression problem. Their strategy aims to regress the rotation matrix and translation vector to estimate the pose in order to avoid the ambiguity caused by the gimbal lock problem that occurs when faces get close to profile view. The authors use a fast lightweight backbone convolutional neural network and apply both a landmark regressor and a cost function with two terms, the weighted parameter distance cost (WPDC) and the vertex distance cost (VDC).  This cost function minimizes the vertex distances between a fitted 3D face and the ground truth. They also establish a 3D aided short-video-synthesis method which helps to achieve smoother estimation results in videos.

\par Occlusion has also been addressed in the problem of HPE, aided by model-based approaches.  
The authors in \cite{ref14} estimate the head pose of partially occluded faces by tracking the displacement of a face feature with respect to the center of the head. They use CamShift to track the center of the head and a iterative Lucas-Kanade optical flow tracker to track the feature face point. This method has disadvantages, since it requires the mouth not to be occluded and also it relies on discontinued software and hardware. \par The authors of \cite{ref16} focus on achieving robust facial landmark detection for severe occlusions and images with large head poses. They use landmark visibility probabilities to measure if a landmark is visible, and perform occlusion prediction. They add a prior occlusion pattern loss to aid the performance of the prediction. This work, however, does not have real-time tracking capabilities and does not specifically focus on estimating poses. \par The method of \cite{ref17} estimates facial landmark locations, head pose and facial deformation under facial occlusions. This procedure updates each estimation parameter based on the previously estimated values of the others. According to the authors, the combined framework achieves better results in head pose estimation than other methods that use all landmarks (as a rigid model) instead of only the ones that are visible. However, this work  only evaluates yaw angles.

\subsection{Learning-based strategies}
\par  As mentioned by the authors of \cite{ref18}, model-based methods rely on the chosen head model and are sensitive to errors in landmark/keypoint detection. Learning-based HPE approaches have been proposed to avoid these drawbacks. The idea behind such strategies is to directly estimate the pose from image inputs to neural networks. These networks are commonly trained through regression of Euler angles or rotation matrix parameters using head pose datasets with ground-truth pose annotations. \par The authors from \cite{ref18} propose a deep learning strategy which employs a backbone neural network  augmented with three fully-connected layers, each one used to predict a different Euler angle. The authors follow a multi-loss approach that combines a classification loss with a weighted regression loss for each angle. Cross-entropy and a mean squared error losses are used for the classification and regression, respectively. The classification component aids the model to predict the vicinity of the pose and the regression component helps it to achieve fine-grained estimation. This multi-loss approach has inspired other authors for the HPE problem \cite{ref1,ref64}.
\par Another solution, FSA-Net \cite{ref19}, applies the soft-stage wise regression problem defined in \cite{ref20} to solve the HPE challenge. Feature maps from input images are extracted and fused together across several stages. Each stage outputs probabilities distributions for the angle interval classes and refines the estimation within an angle interval assigned by the previous stage. The estimated pose is given by the soft-stage regression function, which corresponds to the sum of the product between probability distribution and the values of pose groups at each stage. 
\par The method \textit{img2pose} presented in \cite{ref21} proposes a novel real-time capable solution to simultaneously perform face detection and head pose estimation with 6 degrees of freedom (Euler rotation and 3D translation vectors) in an image without requiring a prior face detection step. This estimation is computationally less costly than the one of model-based approaches which regress 68x2D=136 elements, instead of only 6. Moreover, this pose allows to align the 3D face with its location in an image, which eliminates the need for face detectors. 
\par The authors of \cite{ref1} extend the multi-loss approach of Hopenet \cite{ref18} for full $360^\circ$ yaw estimation. They generate a new dataset with full range of yaws by combining 300W-LP with computed Euler angle data from the CMU Panoptic Dataset. They use binary-cross entropy as for the classification loss and introduce a new wrapped loss for the regression component. They also utilize a lighter backbone network to facilitate real-time applications. The modifications made to Hopenet achieved SoTA of performance for full-range head pose estimation. 

\par In \cite{quatnet}, a quaternion-based multi-regression loss method (QuatNet) is proposed. The model is trained exclusively with RGB images. The loss combines L2 regression loss for precise angle prediction and ordinal regression loss to learn robust features that rank different intervals of angles. This allows to address the non-stationary property in head pose estimation, where pose variations within each angle interval is different.
\par The ambiguity problem of rotation labels is addressed in \cite{6drepnet}. They propose a landmark-free method (6DrepNet) that uses rotation matrix representation and a geodesic loss for efficient and robust head pose regression. Besides removing ambiguity problems, the authors claim it avoids performance stabilizing measures such as the discretization of the rotation variables into classification problems. 
\par The authors from \cite{lightweight} claim that landmark-free methods disregard the perspective distortion in face images, caused by the misalignment of the face with the camera coordinate system. They propose an image rectification method to reduce this effect and to achieve higher HPE accuracy with a lightweight network. However, this method involves a face detection step and  does not take into account  the impact of occlusions when rectifying a head image.



\subsection{Relations with SoTA methodologies}

In this section we define and fit our methodology within the hierarchical representations for HPE methods presented in the extensive survey of \cite{ref44}. We therefore frame our technique according to the following core aspects:

\begin{itemize}
  \item \textbf{Head pose representation} - In this work, we focus on estimating the rotations that define the head pose. There are diverse ways of representing these rotations, in particular, with (i) {\em quaternions}~\cite{ref51}, with rotation matrices~\cite{6drepnet,ref53}, or with (ii) {\em  Euler angles}~\cite{ref1,ref18,ref19}. Our technique represents head pose according to {\em Euler angles} ($\alpha, \beta, \gamma$ - yaw, pitch and roll), which can be visually interpreted.
  
  \item \textbf{Type of input data} - The training and testing datasets depend on the type of input data used. HPE can be performed in three different types of data: (i) {\em Depth images}~\cite{ref56,ref25}, which include RGB and depth information of images; (ii) {\em Videos} ~\cite{ref57,ref58}, that is, sequences of annotated frames for HPE with tracking purposes; and (iii) {\em 2D RGB images}~\cite{ref24,ref26}. Our method performs HPE with {\em 2D RGB image inputs}, since it is more robust to solve in-the-wild HPE problems \cite{ref44}.
  \item \textbf{Pre-processing techniques} - There are three preprocessing techniques which are common in HPE problems: (i) {\em Landmark detection}~\cite{ref59}; (ii) {\em 3D modeling}~\cite{ref60}; and (iii) {\em face detection} \cite{ref61}. Since our method estimates the pose directly from 2D image intensities instead of using landmarks or 3D models, we apply a {\em face detection} technique to find the head region.
  \item \textbf{HPE techniques} - The survey of~\cite{ref44} splits 2D RGB HPE techniques in (i) {\em training}~\cite{ref1,ref18,ref66} and (ii) {\em training-free}~\cite{ref62,ref63} frameworks. Our method fits as a {\em training} technique based on convolutional neural networks.
\end{itemize}

\subsection{Summary and outline}

From the above literature review, we find that the challenge of occlusion in head pose estimation is still scarce. For instance, the system in \cite{ref14} requires the mouth not to be occluded. The procedure in \cite{ref16} although addressing occlusions, it only focus  on landmark detection and it is not extended for real-time tracking, and while the method in \cite{ref17} includes pose estimation, it only evaluates yaw angles and displays low accuracy for large yaw values. \par
Model-based methods rely on the chosen head/face model and are very sensitive to landmark detection and tracking errors. They are also more susceptible both to self-occlusions (extreme poses for example) and object occlusions. Learning-based methods do not require the detection of landmarks and therefore avoid the occlusion problem mentioned above, while outperforming model-based methods. \par Consequently, our work follows a learning-based approach based on the framework from \cite{ref18},  but adapted to facial occlusions.  Specifically, we introduce a novel latent space regression component to o approximate the latent space representation of occluded images to that of non-occluded images. Results show that this aids the model to generalize better to occlusions without deprecating the performance in non-occluded faces.  Furthermore, since our approach requires occluded labeled data to train and test the neural network, we also propose a novel procedure to generate synthetic occlusions in existing head pose labeled datasets. 

\par The outline of this paper is as follows: Section \ref{section: generating_occl} describes the synthetic occlusion generation procedure and the utilized head pose datasets; Section~\ref{sec: methodology} outlines our end-to-end learning methodology for HPE under occlusions. In Section~\ref{sec: evaluation} we  compare our implementation to other SoTA methodologies in three benchmark datasets:  The first two, BIWI \cite{ref25} and AFLW2000 \cite{ref24}, with synthetically occluded and original non-occluded images; and the third, Pandora \cite{ref55}, with real-life, natural occlusions.  We perform ablation studies for the hyper-parameters used within our framework, as well as for different occlusions severities. We visualize our model's latent space representation and compare it to a SoTA. Also, we test our model in the real occlusion case concerning an autonomous feeding robot. Lastly, in Section~\ref{sec: conclusion} we make concluding remarks and mention some of the current limitations and future work to further improve our methodology.

\section{Generating a synthetic occluded dataset}
\label{section: generating_occl}

In this section, we introduce a procedure capable of generating synthetic occlusions in images, and describe the datasets used for the head pose estimation.

\subsection{Synthetic occlusion generation procedure}
\label{section: syntoccl}


The generation of synthetic data for training in deep learning frameworks has become ever more common and has been proven to be essential in the enlargement of training sets and in improving the generalization and accuracy of learning models {\em e.g.}, medical segmentation~\cite{ref46}, autonomous driving~\cite{ref47} and pose estimation~\cite{ref45,ref64}. 
\par 
We use existing 2D image head pose datasets that contain thousands of images and respective ground truth pose annotations. However, we intend to train and test our model for occluded images and most datasets contain few or no face occlusions. Thus, we propose a new  auxiliary  pipeline to generate synthetic occlusions for any  RGB  image and develop the new occluded datasets required for the training and testing of the deep neural networks implemented in our framework.

\begin{figure}[] 
    \centering
  \subfloat[3-step procedure for the first frame]{%
       \includegraphics[clip,width=1.0\columnwidth]{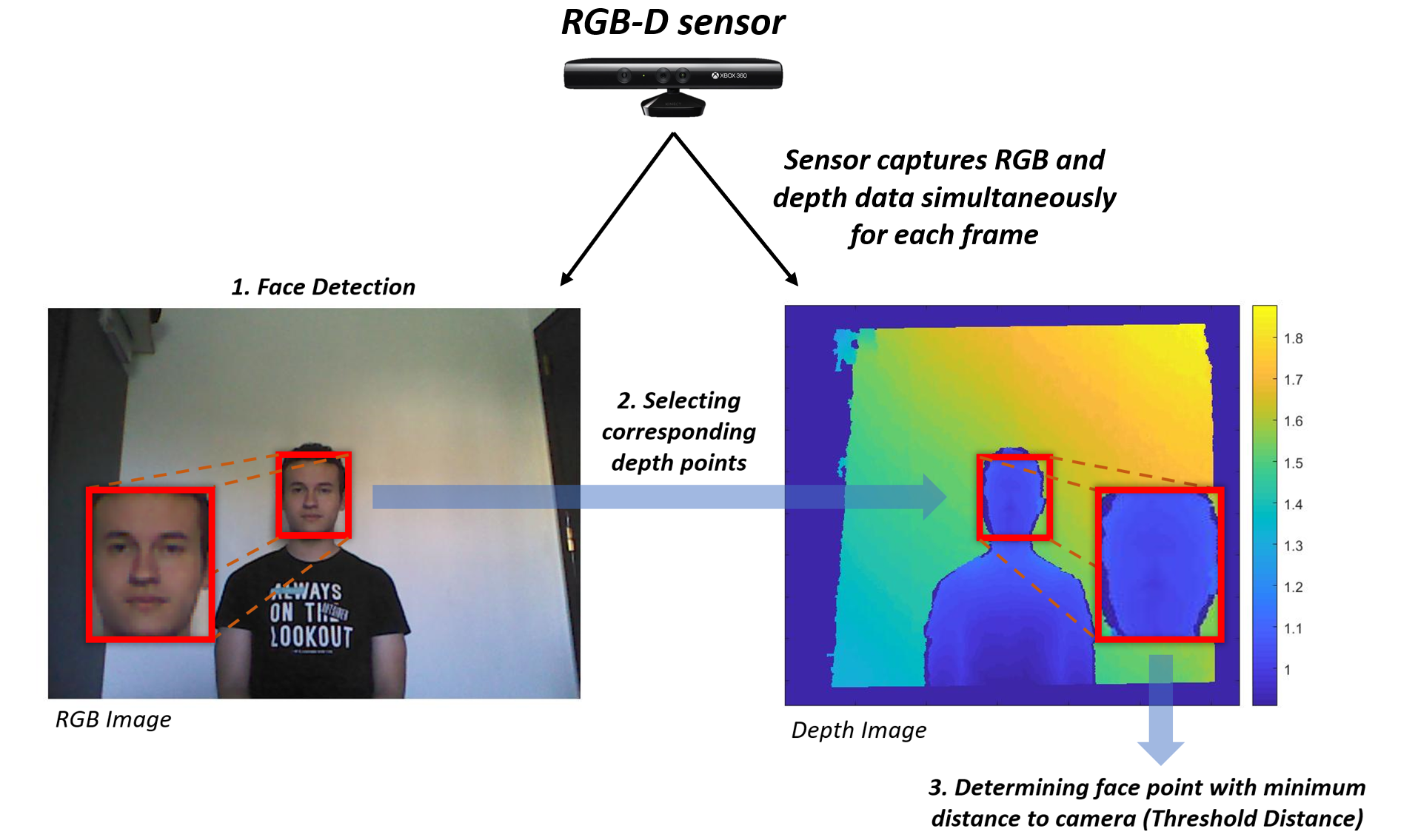}}
    \hfill
  \subfloat[Using threshold distance to extract occlusions]{%
        \includegraphics[clip,width=1\columnwidth]{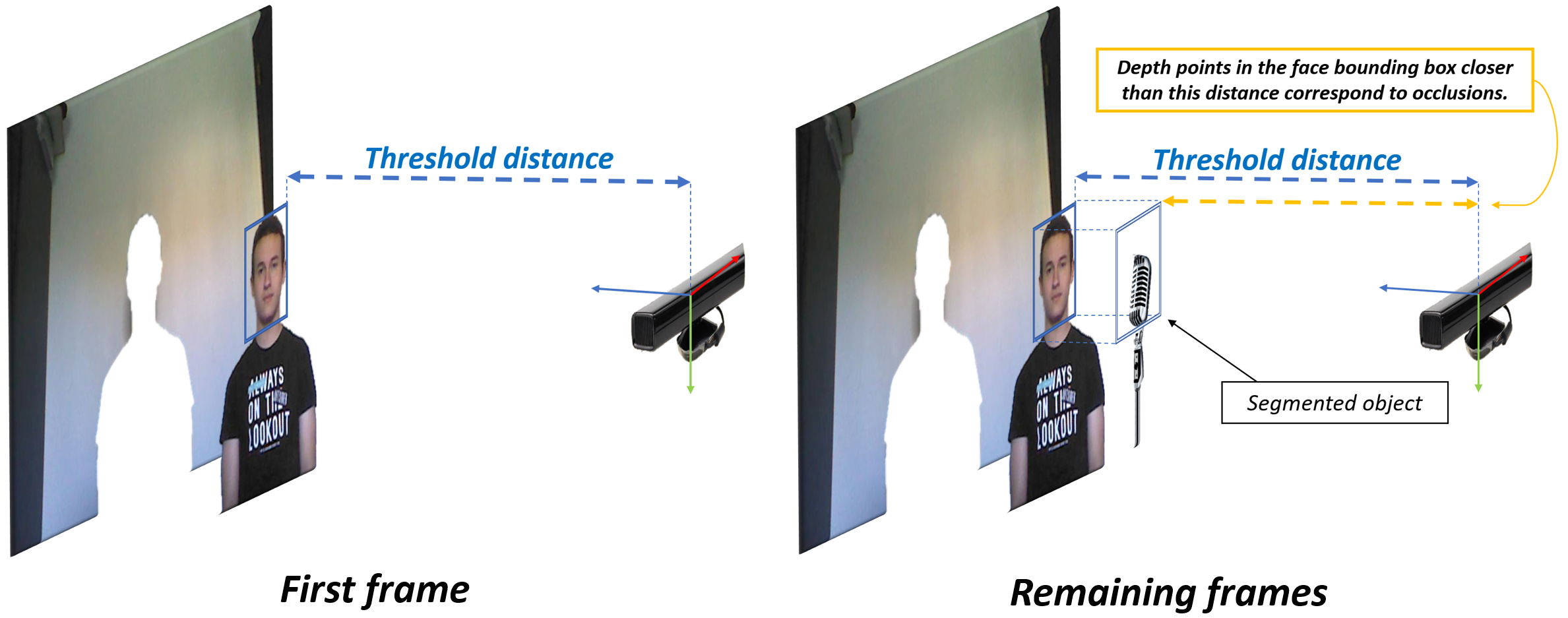}}
  \caption{ Determining threshold distance for occlusion segmentation (first frame only). The idea is to use the first recorded frame to find the distance between the camera and the face using the 3-step procedure in (a): 1. Apply a face detector to the RGB image of the frame to obtain the face bounding box; 2. Select the depth image points within that bounding box; 3. Determine the minimum distance point to the camera, the threshold distance illustrated in (b). With this, it is possible to determine which pixels in the face region correspond to occlusions that we desire to extract: these points will have a smaller depth when compared to the threshold depth.}
  \label{fig: firstframe} 
\end{figure}

\begin{figure}[h]
 \centering
 \includegraphics[width=1.0\textwidth]{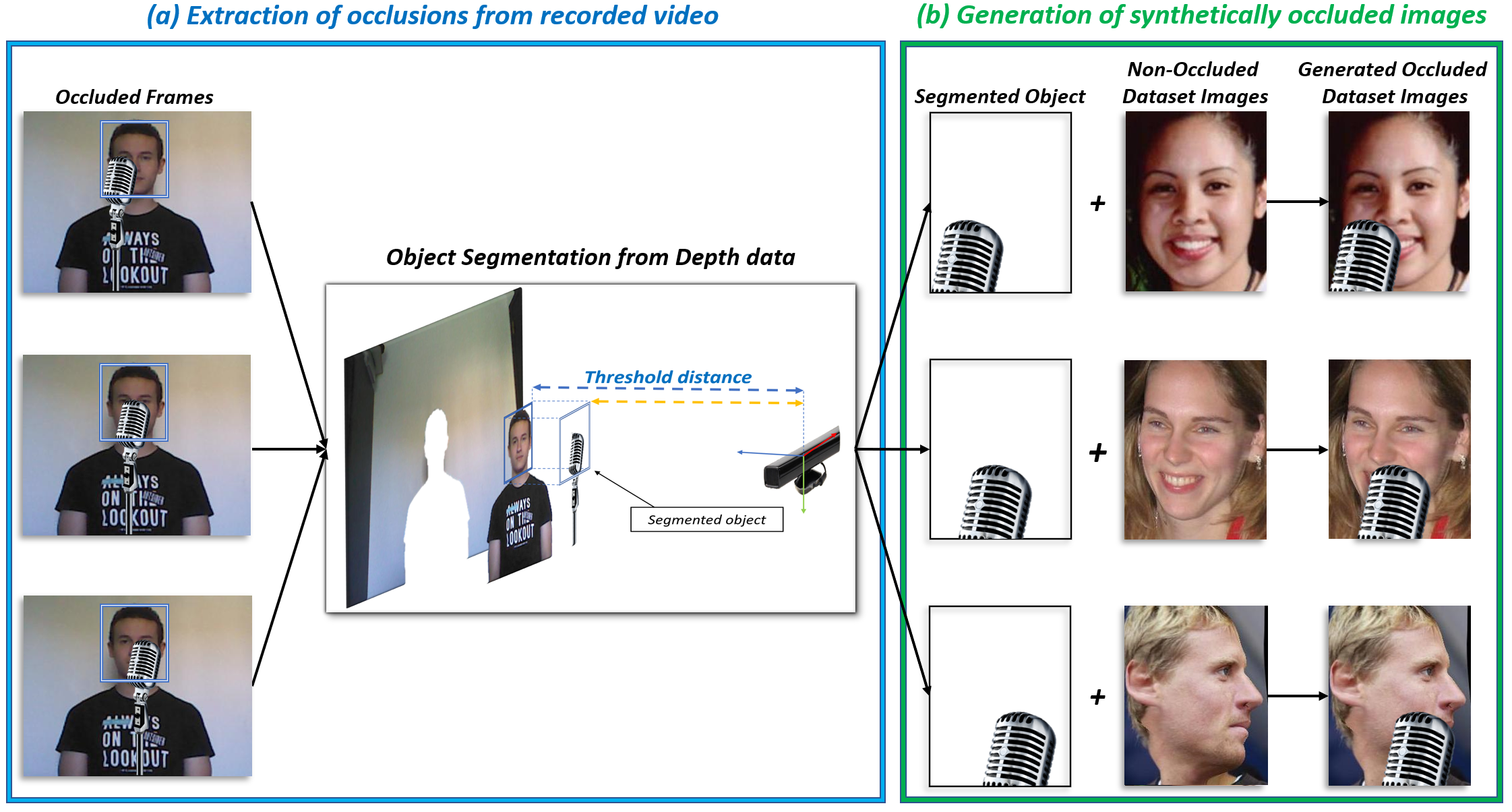}
 \caption{ Overall procedure for synthetic occlusion generation in non-occluded images. }
 \label{fig: occlusiongeneration}
\end{figure}

\par Our procedure to generate synthetic occlusions in images is based on the use of  2D RGB depth data  (camera distance to an object). To that end, we use RGB-D cameras which combine RGB and depth sensors and are capable of simultaneously recording the necessary data. We use a Kinect for that purpose. Since such sensors may have inaccuracies, we implement DBSCAN, a density-based clustering algorithm, to remove outliers from the depth data. The procedure, illustrated in Fig.~\ref{fig: occlusiongeneration}, is as follows:

\begin{enumerate}
  \item Record a in-house video sequence using an RGB-D camera and extract occlusions (Fig.~\ref{fig: occlusiongeneration} (a)):
  \begin{enumerate}
      \item The first frame must be occlusion free (Fig. 2 (a)). We detect the face in the RGB image, select the corresponding pixels in the depth image and find the one of lowest depth (closest to the camera). This point will correspond to the threshold distance.
      \item The remaining frames are occluded. For each one of them we check the depth image for points in the face bounding region at a smaller distance than the threshold depth. Those points belong to occlusion objects (Fig. 2 (b)). We extract the corresponding RGB pixels to obtain and store the segmented occlusion object. 
  \end{enumerate}
  \item When the occlusions from every frame are stored, we re-scale and superimpose the occlusions in the non-occluded face images of benchmark datasets. This way, we generate new occluded labeled datasets. (Fig. \ref{fig: occlusiongeneration} (b))
\end{enumerate}

\par This way, it is straightforward to plug any objects' occlusions in benchmark datasets in a realistic fashion.  Instead of superimposing generic synthetic templates on faces, this method allows the generation of any occlusion that appears in specific scenarios/contexts, similar to how they occur in real life. 
Notice that this procedure only requires depth data while recording the in-house video sequence and can be applied to any benchmark dataset of RGB images.

\subsection{Head pose datasets}

Following \cite{ref44}, three types of input data can be used: 2D images, depth images and video sequences. We focus on head pose estimation for 2D images, which are ideal for the most challenging in the wild problems. Three head pose datasets are used to apply our procedure and generate new occluded datasets: 300W-LP \cite{ref24}, BIWI \cite{ref25} and AFLW2000 \cite{ref24}.
\par
The 300W-LP dataset consists of 61,225 face samples and respective vertically flipped versions for a total over 122K examples. It covers a large variation of identity, expression, illumination conditions, pose and it provides facial landmark annotations from which it is possible to extract the pose of the head. It is commonly used in the training process of head pose estimation works \cite{ref18, ref1}. The BIWI dataset contains over 15K images of 20 people and covers about $\pm 75^\circ$ yaw and $\pm 60^\circ$ pitch. It is one of the most commonly used benchmarked datasets for HPE challenge. For each frame, it provides a depth image, the corresponding RGB image (both 640x480 pixels), and the ground truth pose annotation. AFLW2000 is a dataset that contains 2K real-world images of diverse head poses which are challenging for face detectors. The pose images include varying lighting and background conditions. This dataset contains annotations for 68 image-level facial landmarks from which the pose can be extracted.  
\par We use occluded versions of the 300W-LP dataset in training and test our methodologies in occluded and non-occluded versions of the BIWI and AFLW2000 datasets.
\par  We also test our method with real-life, natural occlusions from Pandora \cite{ref55}, a dataset for head and shoulder pose, as well as head center localization. Inspired by the automotive context, it simulates driving poses from the point of view of a camera placed inside a dashboard. It contains head and shoulder ground truth pose annotations expressed as yaw, pitch and roll (head pose ranges are $\pm 70^\circ$  roll, $\pm 100^\circ$  pitch and $\pm 125^\circ$  yaw). It includes more than 250K RGB (1920x1080 pixels) and depth images (512x424) with the corresponding annotation. Overall, 10 male and 12 female actors were recorded five times with a Microsoft Kinect One. For testing purposes, we utilize 9619 head occluded images where actors wear garments such as sun glasses, scarves, caps and masks.



\section{Methodology for head pose estimation with occlusions}
\label{sec: methodology}

Next, we detail our methodology for the HPE challenge, giving particular attention on how the components of the overall loss are computed. 

\subsection{End-to-end Multi-loss Approach With Latent Space Regression}
\label{sec: latent}

\begin{figure}[h]
 \centering
 \includegraphics[width=1.0\textwidth]{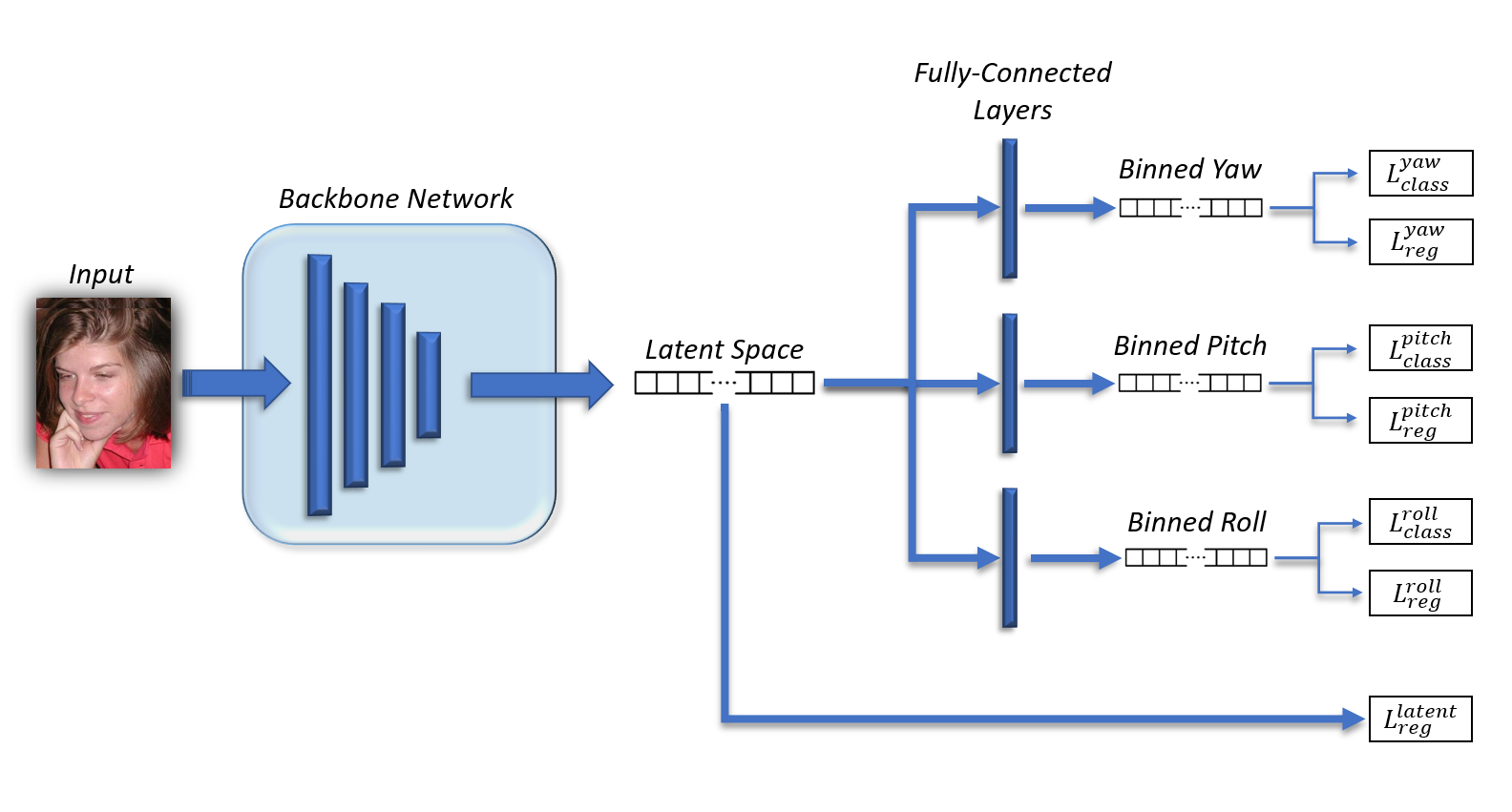}
 \caption{Proposed multi-loss with latent space regression for HPE estimation.}
 \label{fig:approach1}
\end{figure}

Inspired in \cite{ref18}, we propose a new methodology tailored to deal with occlusion as illustrated in Fig.~\ref{fig:approach1}. A 2D RGB image is the input of any backbone network. This backbone network is expanded with three extra fully-connected layers which will be used to output the predictions for each Euler angle. The output of the final layer in the backbone network is flattened into a vector which becomes the input for each fully-connected layer. The output of each of the three layers will be a vector of logits.
Each vector contains the raw prediction scores (real numbers in range [$-\infty$, $+\infty$]) for the predicted angle. Each angle in the vector, is represented as a bin that covers a small range in degrees ({\it e.g.}, [$0^\circ$, $3^\circ$], [$3^\circ$, $6^\circ$], ...). The size of these vectors depends on both the angle interval/span for each bin, and the full prediction range for the given Euler angle ({\it e.g.}, predicting yaw angles in the range [$-90^\circ$, $90^\circ$]). Also, in Fig.~\ref{fig:approach1}, it can be seen that the output of each fully-connected layers is used in a multi-loss scheme that comprises the combination of a classification and regression component to provide an overall loss for a given Euler angle. For the classification task, a softmax activation function plus a cross-entropy loss (also known as categorical cross-entropy loss or softmax loss) is applied to the n-dimensional vector output of the fully-connected layer. The softmax function converts logits into probabilities by computing the exponents of each bin output and normalizing it by the sum of those exponents so that all probabilities in the activated vector add up to one:




\small
\begin{equation}
S(y_i) = \frac{e^{y_i}}{\sum_{j = 1}^{n}e^{y_j}}
\label{eq:softmax}
\end{equation}
\normalsize
where $y_i$ is the logit for class $i$. Afterwards, the cross-entropy loss result is computed by equation (\ref{eq:crossentropy}), where $t_i$ and $S(y_i)$ are the ground-truth (0 or 1) and the activation result of the score for each of the $C$ angle classes/bins, respectively.
\small
\begin{equation}
L_{class} = - \sum_{i}^{C}t_i log(S(y_i))
\label{eq:crossentropy}
\end{equation}
\normalsize

In addition to the classification loss, the regression component is introduced to determine and regress the error between the predicted angle and the ground truth in degrees. It is possible to determine the predicted angle in degrees by using the bin probabilities obtained from softmax activation to calculate the expectation of the given angle:

\small
\begin{equation}
\theta _{pred} = w \sum_{i = 1}^{N} p_i (i-\frac{1+N}{2})
\label{eq:expectedvalue}
\end{equation}
\normalsize
where $\theta _{pred}$ is the predicted angle in degrees, $w$ is the width of the bin in degrees, $N$ is the number of bins for classification, and $p_i$ is the probability of the angle belonging to bin $i$. The offset $\frac{1+N}{2}$ shifts the bin indices to the respective bin centres, as mentioned in \cite{ref1}. The loss used for the regression component is the mean squared error (MSE) between the predicted angle $\theta _{pred}$ and the ground truth angle $\theta _{gt}$, for $N$ predictions.

\small
\begin{equation}
L_{reg}  = \frac{1}{N}\sum_{i = 1}^{N}(\theta _{pred}-\theta _{gt})^2
\label{eq:lossreg}
\end{equation}
\normalsize

The classification component (\ref{eq:crossentropy}) aims to help the model to predict the vicinity of each pose angle by classifying it in an angle interval bin. The regression component (\ref{eq:lossreg}) is then introduced to aid the model in achieving fine-grained angle predictions.


\par The classification and regression loss for the Euler angles are combined using a regularization hyper-parameter $\alpha$ to manage the trade-off between the two terms. Overall, three losses are used to train the Euler angles:

\small
\begin{equation}
\centering
\begin{split}
L_{yaw} = L_{class}^{yaw}(y,\hat{y}) + \alpha \, L_{reg}^{yaw}(y, \hat{y}) \\
L_{pitch} = L_{class}^{pitch}(y, \hat{y}) + \alpha \, L_{reg}^{pitch}(y,\hat{y}) \\
L_{roll} = L_{class}^{roll}(y, \hat{y}) + \alpha \, L_{reg}^{roll}(y,\hat{y}) \\
\end{split}
\label{eq:losses}
\end{equation}
\normalsize

\noindent
where $y$ is the predicted value and $\hat{y}$ is the ground truth for the respective Euler angle loss, provided in the training dataset.


\subsubsection{Introducing latent space regression}
\par
We introduce a new latent space regression loss, crucial to aid the model to generalize better to occlusions. The latent space, also known as embedding space, is the abstract low-dimensional space that contains the highest-level feature values of the neural network. These values encode the most relevant inner representation of the observed input data and allow to semantically place similar (pose) outputs closer in the latent space. The idea behind this loss is to approximate the latent space representation of occluded images to that of non-occluded images and therefore allow our model to better predict poses with occlusions. The additional latent space regression loss is the MSE between the predicted latent embedding for occluded images $E_{pred}$ and the ground truth embedding for non-occluded images $E_{gt}$. Thus, for N predictions we have:

\small
\begin{equation}
L_{latent}  = \frac{1}{N}\sum_{i = 1}^{N}(E_{pred}-E_{gt})^2
\label{eq:latentloss}
\end{equation}
\normalsize

Note that the ground truth latent space embeddings $E_{gt}$ for non-occluded images is previously extracted from inference on the Hopenet network. 
The overall loss is the sum of angle and latent losses with a hyper-parameter $\beta$ to handle the trade-off between them: 
\small
\begin{equation}
\centering
\begin{split}
L = (1-\beta) (L_{yaw} + L_{pitch} + L_{roll}) + \beta L_{latent}
\end{split}
\label{eq:losstotallatent}
\end{equation}
\normalsize


\begin{figure}[h]
 \centering
 \includegraphics[width=1.0\textwidth]{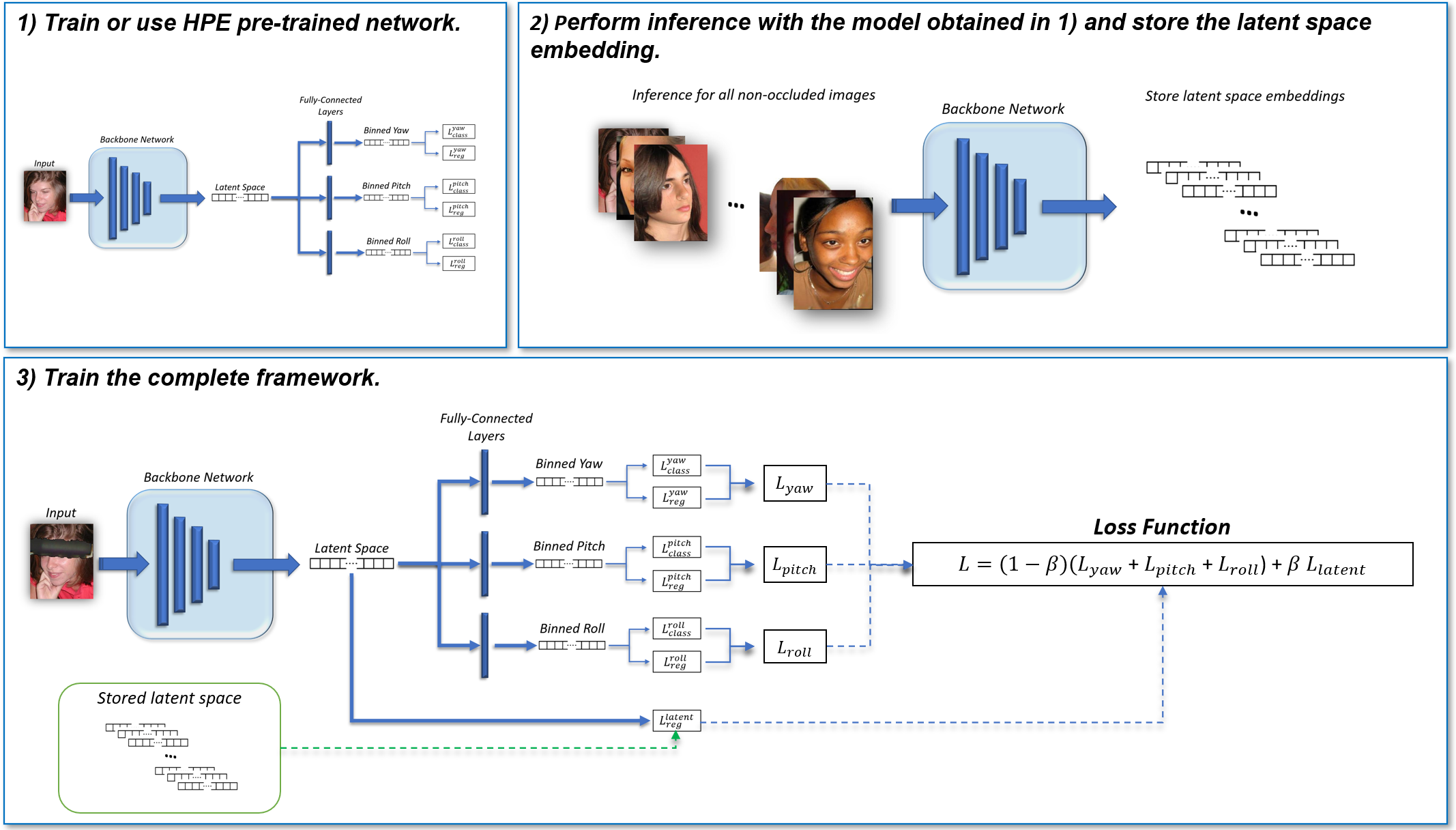}
 \caption{Training procedure representation.}
 \label{fig:training_procedure}
\end{figure}

Our complete training procedure is illustrated in Fig. \ref{fig:training_procedure} and comprises the following steps:

\begin{enumerate}
  \item \label{1} Firstly, we either (i) train or (ii) use a pre-trained model for head pose estimation in non-occluded images. The framework is the same as shown in Fig.~\ref{fig:approach1} apart from the latent space loss;
  \item We perform inference with the model obtained in \ref{1}), for each non-occluded image and store the flattened output of the final layer in the backbone network. This output corresponds to the representation of the latent space and is used as ground truth for the train with occluded images; 
  \item Finally, we use the occluded dataset and train the complete framework of Fig.~\ref{fig:approach1}, where $L_{class}^a$ and $L_{reg}^a$ are the cross-entropy classification loss and MSE regression loss for Euler angle $a$ (yaw, pitch or roll), and $L_{reg}^{latent}$ is the MSE regression loss for the latent space.
\end{enumerate}

\section{Results and discussion}
\label{sec: evaluation}

In this section we describe the experiments of the proposed framework. Concretely, we  (i) study the impact of varying the hyper-parameters of the framework, (ii) perform a comparison of related SoTA methodologies and (iii) perform an ablation study regarding the severity of the occlusion. 

\subsection{Multi-loss head pose estimation with latent space regression}

We evaluate the methodology on both the original and synthetically occluded versions of the BIWI and AFLW2000 datasets as well as with natural occlusions from the Pandora dataset. 
\par We train the framework defined in Section \ref{sec: latent} using  synthetically occluded and non-occluded images of 300W-LP  and ResNet-50 as the backbone network. The face images are cropped to the pre-defined input dimension of the ResNet-50 network, 224x224 pixels, and the mean and standard deviation of ImageNet is used to normalize the data. We use 66 bins for classification, each bin has a width of $3^\circ$. Our classification vectors cover a range from $-99^\circ$ to $99^\circ$ for each Euler angle. There are 31 images of the AFLW2000 dataset that are not used in testing since their pose angles surpass this range.

\subsubsection{$\alpha$-parameter ablation study}

\begin{table}[h]
\begin{center}
\caption{Head pose estimation MAE ($^\circ$) tests with BIWI for different angle regression weights ($\alpha$).}\label{table:alpha_biwi}
\vspace{0.5cm}
\resizebox{0.7\columnwidth}{!}{
\begin{tabular}{c|cccc|c|c|l|l}
\textit{BIWI \cite{ref25}}                                                 & \multicolumn{4}{c|}{Occluded Images}                                        & \multicolumn{4}{c}{}                                                                   \\ \hline
\begin{tabular}[c]{@{}c@{}}$\alpha$\end{tabular} & yaw   & pitch & roll  & \begin{tabular}[c]{@{}c@{}}Average \\ MAE\end{tabular} & \multicolumn{4}{c}{\begin{tabular}[c]{@{}c@{}}Combined \\ Average\end{tabular}}           \\ \hline
0.5                                                           & 5.250 & 6.529 & 4.255 & 5.345                                               & $\alpha$                            & \multicolumn{3}{c}{MAE}                             \\ \cline{6-9} 
1                                                             & 5.333 & 6.916 & 4.036 & 5.495                                               & \multirow{2}{*}{0.5}             & \multicolumn{3}{c}{\multirow{2}{*}{4.763}}          \\
2                                                             & 5.110 & 6.832 & 3.629 & \textbf{5.190}                                      &                                  & \multicolumn{3}{c}{}                                \\ \cline{6-9} 
5                                                             & 5.477 & 6.542 & 4.304 & 5.441                                               & \multirow{2}{*}{1}               & \multicolumn{3}{c}{\multirow{2}{*}{4.950}}          \\
10                                                            & 5.218 & 7.565 & 4.344 & 5.709                                               &                                  & \multicolumn{3}{c}{}                                \\ \hline
                                                              & \multicolumn{4}{c|}{Non-Occluded Images}                                    & \multirow{2}{*}{2}               & \multicolumn{3}{c}{\multirow{2}{*}{\textbf{4.617}}} \\ \cline{1-5}
0.5                                                           & 4.259 & 4.704 & 3.580 & 5.441                                               &                                  & \multicolumn{3}{c}{}                                \\ \cline{6-9} 
1                                                             & 4.765 & 4.493 & 3.956 & 5.768                                               & \multirow{2}{*}{5}               & \multicolumn{3}{c}{\multirow{2}{*}{4.723}}          \\
2                                                             & 4.242 & 4.041 & 3.845 & 4.043                                               &                                  & \multicolumn{3}{c}{}                                \\ \cline{6-9} 
5                                                             & 4.474 & 4.043 & 3.494 & \textbf{4.004}                                      & \multirow{2}{*}{10}              & \multicolumn{3}{c}{\multirow{2}{*}{4.909}}          \\
10                                                            & 4.196 & 4.503 & 3.628 & 4.109                                               &                                  & \multicolumn{3}{c}{}                               
\end{tabular}
}
\end{center}
\end{table}

We train the framework for 25 epochs with 5 different $\alpha$ parameter values to determine the best trade off for pose estimation training (see (\ref{eq:losses})). The head pose estimation error (MAE) results in synthetically occluded and non-occluded datasets are listed in Tables~\ref{table:alpha_biwi} and~\ref{table:alpha_alfw2000}, for the BIWI and AFLW2000 datasets, respectively. We can observe that, generally, $\alpha = 2$ produces the smallest average MAE errors. In particular for occluded images, the lowest MAE errors across all datasets correspond to the networks trained with that parameter value. We can also observe that the largest errors tend to occur for $\alpha=1$. This result highlights the importance of performing this study and selecting the ideal value for this parameter, since $\alpha=1$ corresponds to both losses having the same weight. For the following tests, we use $\alpha= 2$ as the weight for the head pose multi-loss framework. 




\begin{table}[h]
\begin{center}
\caption{Head pose estimation MAE ($^\circ$) tests with AFLW2000 for different angle regression weights ($\alpha$).}\label{table:alpha_alfw2000}
\vspace{0.5cm}
\resizebox{0.7\columnwidth}{!}{
\begin{tabular}{c|cccc|c|c|l|l}
\textit{AFLW2000 \cite{ref24}}                                             & \multicolumn{4}{c|}{Occluded Images}                                         & \multicolumn{4}{c}{}                                                                   \\ \hline
\begin{tabular}[c]{@{}c@{}}$\alpha$\end{tabular} & yaw    & pitch & roll  & \begin{tabular}[c]{@{}c@{}}Average \\ MAE\end{tabular} & \multicolumn{4}{c}{\begin{tabular}[c]{@{}c@{}}Combined \\ Average \end{tabular}}           \\ \hline
0.5                                                           & 6.227  & 8.271 & 5.713 & 6.737                                               & $\alpha$                            & \multicolumn{3}{c}{MAE}                             \\ \cline{6-9} 
1                                                             & 6.411  & 8.713 & 6.017 & 7.047                                               & \multirow{2}{*}{0.5}             & \multicolumn{3}{c}{\multirow{2}{*}{6.089}}          \\
2                                                             & 5.672  & 8.101 & 5.783 & \textbf{6.519}                                      &                                  & \multicolumn{3}{c}{}                                \\ \cline{6-9} 
5                                                             & 6.156  & 8.279 & 5.841 & 6.759                                               & \multirow{2}{*}{1}               & \multicolumn{3}{c}{\multirow{2}{*}{6.4075}}         \\
10                                                            & 5.4044 & 8.407 & 5.923 & 6.578                                               &                                  & \multicolumn{3}{c}{}                                \\ \hline
                                                              & \multicolumn{4}{c|}{Non-Occluded Images}                                     & \multirow{2}{*}{2}               & \multicolumn{3}{c}{\multirow{2}{*}{\textbf{5.954}}} \\ \cline{1-5}
0.5                                                           & 5.281  & 6.544 & 4.497 & 5.441                                               &                                  & \multicolumn{3}{c}{}                                \\ \cline{6-9} 
1                                                             & 5.675  & 6.868 & 4.760 & 5.768                                               & \multirow{2}{*}{5}               & \multicolumn{3}{c}{\multirow{2}{*}{6.1065}}         \\
2                                                             & 4.886  & 6.636 & 4.643 & \textbf{5.389}                                      &                                  & \multicolumn{3}{c}{}                                \\ \cline{6-9} 
5                                                             & 5.403  & 6.413 & 4.546 & 5.454                                               & \multirow{2}{*}{10}              & \multicolumn{3}{c}{\multirow{2}{*}{6.0185}}         \\
10                                                            & 4.986  & 6.695 & 4.696 & 5.459                                               &                                  & \multicolumn{3}{c}{}                               
\end{tabular}
}
\end{center}
\end{table}

\subsubsection{Latent Space Regression Weight Ablation Study}
\label{section 5.1.2}

\begin{table}[]
\begin{center}
\caption{Head pose estimation MAE ($^\circ$) tests with BIWI for different latent space regression weights ($\beta$). LSR stands for latent space regression.}\label{table:beta_biwi}
\vspace{0.5cm}
\resizebox{0.9\columnwidth}{!}{
\begin{tabular}{c|cccc|ccll}
\textit{BIWI \cite{ref25}}               & \multicolumn{4}{c|}{Occluded images}                                                                      &                                                   &                                 \\ \hline
Methods            & Yaw            & Pitch          & Roll           & \begin{tabular}[c]{@{}c@{}}Average \\ MAE\end{tabular} & \multicolumn{2}{c}{\begin{tabular}[c]{@{}c@{}}Combined \\ Average\end{tabular}}     \\ \hline
FSA-Net \cite{ref19}            & 10.987         & 9.848          & 7.846          & 9.560                                                  & \multicolumn{1}{c|}{$\beta$}                         & MAE                             \\ \cline{6-7} 
WHENet \cite{ref1}            & 21.101         & 13.188         & 11.225         & 15.171                                                 & \multicolumn{1}{c|}{\multirow{2}{*}{FSA-Net}}     & \multirow{2}{*}{7.364}          \\
Hopenet \cite{ref18}           & 6.725          & 8.616          & 7.338          & 7.560                                                  & \multicolumn{1}{c|}{}                             &                                 \\ \cline{6-7} 
6DRepNet \cite{6drepnet}                & 7.883          & 14.983         & 9.665          & 10.844                                                 & \multicolumn{1}{c|}{\multirow{2}{*}{WHENet}}      & \multirow{2}{*}{9.822}          \\
DAD-3D \cite{dad3d}             & 5.532          & 7.924          & 7.478          & 6.978                                                  & \multicolumn{1}{c|}{}                             &                                 \\ \cline{6-7} 
Lightweight \cite{lightweight}        & 10.784         & 9.654          & 10.309         & 10.249                                                 & \multicolumn{1}{c|}{\multirow{2}{*}{Hopenet}}     & \multirow{2}{*}{5.661}          \\
LSR ($\beta$ = 0)     & 5.990          & 7.778          & 4.346          & 6.038                                                  & \multicolumn{1}{c|}{}                             &                                 \\ \cline{6-7} 
LSR ($\beta$ = 0.5)   & 5.797          & 7.394          & 4.537          & 5.910                                                  & \multicolumn{1}{c|}{\multirow{2}{*}{6DRepNet}}          & \multirow{2}{*}{7.252}          \\
LSR ($\beta$ = 0.990) & 5.798          & 6.881          & 4.572          & 5.750                                                  & \multicolumn{1}{c|}{}                             &                                 \\ \cline{6-7} 
LSR ($\beta$ = 0.999) & \textbf{5.174} & 6.622          & 4.117          & 5.304                                                  & \multicolumn{1}{c|}{\multirow{2}{*}{DAD}}         & \multirow{2}{*}{5.298}          \\
LSR ($\beta$ = 1)     & 5.429          & \textbf{4.823} & \textbf{3.467} & \textbf{4.573}                                         & \multicolumn{1}{c|}{}                             &                                 \\ \hline
                   & \multicolumn{4}{c|}{Non-occluded images}                                                                  & \multicolumn{1}{c|}{\multirow{2}{*}{Lightweight}} & \multirow{2}{*}{7.292}          \\ \cline{1-5}
FSA-Net \cite{ref19}           & 5.420          & 5.568          & 4.515          & 5.168                                                  & \multicolumn{1}{c|}{}                             &                                 \\ \cline{6-7} 
WHENet \cite{ref1}            & 4.395          & 4.346          & 4.674          & 4.472                                                  & \multicolumn{1}{c|}{\multirow{2}{*}{0}}           & \multirow{2}{*}{5.307}          \\
Hopenet \cite{ref18}           & 4.375          & 3.559          & 3.348          & 3.761                                                  & \multicolumn{1}{c|}{}                             &                                 \\ \cline{6-7} 
6DRepNet \cite{6drepnet}                 & \textbf{3.230} & 4.658          & 3.091          & 3.660                                                  & \multicolumn{1}{c|}{\multirow{2}{*}{0.5}}         & \multirow{2}{*}{5.102}          \\
DAD-3D \cite{dad3d}                & 3.314          & 4.630          & \textbf{3.090} & 3.618                                                  & \multicolumn{1}{c|}{}                             &                                 \\ \cline{6-7} 
Lightweight \cite{lightweight}       & 4.267          & 5.015          & 3.722          & 4.335                                                  & \multicolumn{1}{c|}{\multirow{2}{*}{0.990}}       & \multirow{2}{*}{4.925}          \\
LSR ($\beta$ = 0)     & 4.940          & 4.873          & 3.911          & 4.575                                                  & \multicolumn{1}{c|}{}                             &                                 \\ \cline{6-7} 
LSR ($\beta$ = 0.5)   & 4.413          & 4.910          & 3.556          & 4.293                                                  & \multicolumn{1}{c|}{\multirow{2}{*}{0.999}}       & \multirow{2}{*}{4.669}          \\
LSR ($\beta$ = 0.990  & 4.204          & 4.343          & 3.750          & 4.099                                                  & \multicolumn{1}{c|}{}                             &                                 \\ \cline{6-7} 
LSR ($\beta$ = 0.999) & 4.297          & 4.186          & 3.617          & 4.033                                                  & \multicolumn{1}{c|}{\multirow{2}{*}{1}}           & \multirow{2}{*}{\textbf{4.046}} \\
LSR ($\beta$ = 1)     & 4.291          & \textbf{3.086} & 3.179          & \textbf{3.519}                                         & \multicolumn{1}{c|}{}                             & 
\end{tabular}
}
\end{center}
\end{table}


\begin{table}[]
\begin{center}
\caption{Head pose estimation MAE ($^\circ$) tests with AFLW2000 for different latent space regression weights ($\beta$). LSR stands for latent space regression.}\label{table:beta_aflw2000}
\vspace{0.5cm}
\resizebox{0.9\columnwidth}{!}{
\begin{tabular}{c|cccc|ccll}
\textit{AFLW2000 \cite{ref24}}           & \multicolumn{4}{c|}{Occluded images}                                                                      &                                                   &                                 \\ \hline
Methods            & Yaw            & Pitch          & Roll           & \begin{tabular}[c]{@{}c@{}}Average \\ MAE\end{tabular} & \multicolumn{2}{c}{\begin{tabular}[c]{@{}c@{}}Combined \\ Average\end{tabular}}     \\ \hline
FSA-Net \cite{ref19}            & 13.664         & 10.880         & 10.067         & 11.537                                                 & \multicolumn{1}{c|}{$\beta$}                         & MAE                             \\ \cline{6-7} 
WHENet \cite{ref1}             & 16.515         & 11.669         & 10.267         & 12.817                                                 & \multicolumn{1}{c|}{\multirow{2}{*}{FSA-Net}}     & \multirow{2}{*}{8.590}          \\
Hopenet \cite{ref18}            & 12.438         & 10.277         & 8.586          & 10.434                                                 & \multicolumn{1}{c|}{}                             &                                 \\ \cline{6-7} 
6DRepNet \cite{6drepnet}          & 8.904          & 9.799          & 7.408          & 8.704                                                  & \multicolumn{1}{c|}{\multirow{2}{*}{WHENet}}      & \multirow{2}{*}{9.028}          \\
DAD-3D \cite{dad3d}             & 7.084          & 14.953         & 13.680         & 11.906                                                 & \multicolumn{1}{c|}{}                             &                                 \\ \cline{6-7} 
Lightweight \cite{lightweight}        & 10.771         & 9.278          & 8.335          & 9.461                                                  & \multicolumn{1}{c|}{\multirow{2}{*}{Hopenet}}     & \multirow{2}{*}{7.579}          \\
LSR ($\beta$ = 0)     & 5.057          & 7.120          & 4.961          & 5.713                                                  & \multicolumn{1}{c|}{}                             &                                 \\ \cline{6-7} 
LSR ($\beta$ = 0.5)   & 4.891          & 6.424          & 4.918          & 5.411                                                  & \multicolumn{1}{c|}{\multirow{2}{*}{6DRepNet}}    & \multirow{2}{*}{6.181}          \\
LSR ($\beta$ = 0.990) & \textbf{4.714} & 6.360          & 4.906          & 5.327                                                  & \multicolumn{1}{c|}{}                             &                                 \\ \cline{6-7} 
LSR ($\beta$ = 0.999) & 4.741          & 6.254          & 4.765          & \textbf{5.253}                                         & \multicolumn{1}{c|}{\multirow{2}{*}{DAD-3D}}      & \multirow{2}{*}{7.762}          \\
LSR ($\beta$ = 1)     & 5.117          & \textbf{6.075} & \textbf{4.590} & 5.261                                                  & \multicolumn{1}{c|}{}                             &                                 \\ \hline
                   & \multicolumn{4}{c|}{Non-occluded images}                                                                  & \multicolumn{1}{c|}{\multirow{2}{*}{Lightweight}} & \multirow{2}{*}{6.898}          \\ \cline{1-5}
FSA-Net \cite{ref19}            & 5.109          & 6.462          & 3.356          & 5.642                                                  & \multicolumn{1}{c|}{}                             &                                 \\ \cline{6-7} 
WHENet \cite{ref1}             & 4.475          & 6.222          & 5.017          & 5.238                                                  & \multicolumn{1}{c|}{\multirow{2}{*}{0}}           & \multirow{2}{*}{5.220}          \\
Hopenet \cite{ref18}            & 4.965          & 5.250          & 3.956          & 4.724                                                  & \multicolumn{1}{c|}{}                             &                                 \\ \cline{6-7} 
6DRepNet \cite{6drepnet}           & 3.230          & 4.658          & 3.091          & 3.659                                                  & \multicolumn{1}{c|}{\multirow{2}{*}{0.5}}         & \multirow{2}{*}{4.914}          \\
DAD-3D \cite{dad3d}            & \textbf{3.134} & \textbf{4.630} & \textbf{3.090} & \textbf{3.618}                                         & \multicolumn{1}{c|}{}                             &                                 \\ \cline{6-7} 
Lightweight \cite{lightweight}        & 4.267          & 5.015          & 3.722          & 4.335                                                  & \multicolumn{1}{c|}{\multirow{2}{*}{0.990}}       & \multirow{2}{*}{4.882}          \\
LSR ($\beta$ = 0)     & 4.114          & 6.002          & 4.061          & 4.726                                                  & \multicolumn{1}{c|}{}                             &                                 \\ \cline{6-7} 
LSR ($\beta$ = 0.5)   & 3.855          & 5.447          & 3.947          & 4.416                                                  & \multicolumn{1}{c|}{\multirow{2}{*}{0.999}}       & \multirow{2}{*}{\textbf{4.833}} \\
LSR ($\beta$ = 0.990  & 3.709          & 5.517          & 4.083          & 4.436                                                  & \multicolumn{1}{c|}{}                             &                                 \\ \cline{6-7} 
LSR ($\beta$ = 0.999) & 3.813          & 5.420          & 4.003          & 4.412                                                  & \multicolumn{1}{c|}{\multirow{2}{*}{1}}           & \multirow{2}{*}{4.867}          \\
LSR ($\beta$ = 1)     & 4.258          & 5.272          & 3.888          & 4.473                                                  & \multicolumn{1}{c|}{}                             &                                    
\end{tabular}
}
\end{center}
\end{table}

\par Having determined the best $\alpha$, we trained 5 different head pose estimation networks, one for each different value of $\beta$. This parameter controls the weight of the latent and angle losses (see  (\ref{eq:losstotallatent})). We also perform a comparison between our results and  the results of Hopenet \cite{ref18}, WHENet \cite{ref1}, FSA-Net \cite{ref19}, DAD-3D \cite{dad3d}, 6DRepNet \cite{6drepnet} and Lightweight HPE \cite{lightweight}. We have also considered the model-based method of 3DDFA \cite{ref11}. However, as argued in \cite{ref18,ref1}, this approach is not competitive since on the non-occluded BIWI dataset it provides high values for the estimation error.  Our LSR networks are trained for 25 epochs and their parameters are initialized with a pre-trained model provided by the authors of \cite{ref18} (trained with the original 300W-LP dataset). We use the latent space produced by this pre-trained model in non-occluded inference as ground truth in the latent regression loss. To optimize the parameters we use the Adam optimization algorithm with a learning rate of $10^{-5}$.  We tested our method for all datasets with LSR networks trained on occluded and non-occluded images.
For the networks of SoTA methods we utilized the trained models provided by the authors in their official repositories available online on GitHub. The synthetically occluded datasets for training and testing include 25 different recorded occlusions. Tables \ref{table:beta_biwi} and \ref{table:beta_aflw2000} display the results for BIWI and AFLW2000 datasets, respectively.

We observe that FSA-NET, Lightweight and WHENet produce the highest yaw error and average MAE in occluded images.  A possible explanation for the higher errors in FSA-NET is that this model is based on feature aggregation relying on structural spatial information of the face. This causes the occlusion to affect the predictions.  Occlusions also seem to affect the method of image rectification in the Lightweight framework. 
As for WHENet, the results for occluded images are particularly worse on BIWI. This highlights that this method struggles with occlusions specially in low resolution images. Despite the similarities to Hopenet, WHENet's training for $360^\circ$ yaw degrees may lead the model to confuse occlusions with more extreme yaw values where the face is also not as visible.  As for DAD-3D and 6DRepNet, they present the lowest non-occluded estimation errors, but pitch and roll estimations are heavily affected by occlusions.

When we do not consider the latent space regression ($\beta = 0$), similar to \cite{ref18}, the error for occluded images decreases but the performance for non-occluded images deteriorates significantly. This is more evident on the BIWI dataset, where the average MAE for non-occluded images increases by nearly $1^\circ$.  We can also observe that as the $\beta$ parameter increases, the MAE becomes lower for both occluded images and non-occluded images. On AFLW2000 the non-occluded estimation results improve on the Hopenet scores. This confirms that introducing the latent regression loss (see (\ref{eq:losstotallatent})) helps not only to achieve improved generalization for occlusions, but also to avoid detouring from accurate non-occluded pose estimation. While for $\beta=1$ the average MAE errors are the best on BIWI, we verify that this parameter leads to a worse estimation of yaw values, which is the most varied and relevant Euler angle in head pose estimation. From the above, $\beta = 0.999$ provides the best trade-off for the MAE and yaw estimation. 
\begin{table}[!h]
\begin{center}
\caption{Head pose estimation MAE ($^\circ$) tests with Pandora for different latent space regression weights ($\beta$). LSR stands for latent space regression.}\label{table:beta_pandora}
\vspace{0.5cm}
\resizebox{0.72\columnwidth}{!}{
\begin{tabular}{c|cccc}
\textit{Pandora \cite{ref55}}   & \multicolumn{4}{c}{}                                                                                      \\ \hline
Methods            & Yaw            & Pitch          & Roll           & \begin{tabular}[c]{@{}c@{}}Average \\ MAE\end{tabular} \\ \hline
FSA-Net \cite{ref19}            & 11.736         & 8.607          & 6.691          & 9.011                                                  \\
WHENet \cite{ref1}             & 11.352         & 6.281          & 6.453          & 8.029                                                  \\
Hopenet \cite{ref18}            & 10.442         & 7.239          & 6.925          & 8.202                                                  \\
6DRepNet \cite{6drepnet}           & 10.896         & 6.897          & 7.500          & 8.431                                                  \\
DAD-3D \cite{dad3d}             & 9.348          & 7.437          & 8.474          & 8.420                                                  \\
Lightweight \cite{lightweight}        & 10.133         & 8.690          & 7.064          & 8.629                                                  \\
LSR ($\beta$ = 0)     & 9.178          & 6.296          & 6.895          & 7.456                                                  \\
LSR ($\beta$ = 0.5)   & 9.135          & 5.968          & 6.665          & 7.256                                                  \\
LSR ($\beta$ = 0.990) & \textbf{9.084} & 5.690          & 6.565          & 7.113                                                  \\
LSR ($\beta$ = 0.999) & 9.096          & \textbf{5.657} & \textbf{6.215} & \textbf{6.989}                                         \\
LSR ($\beta$ = 1)     & 9.60596        & 5.820          & 6.368          & 7.265 
\end{tabular}
}
\end{center}
\end{table}

 Overall, our lowest average MAE model on BIWI is able to decrease the error for occluded images in approximately $2.4^\circ$, when compared to the best SoTA result (DAD-3D). As for AFLW, our model decreases the error in nearly $3.5^\circ$ in comparison to the best SoTA result (6DRepNet).

 \par Table \ref{table:beta_pandora} display the results for naturally occluded images of the Pandora dataset. FSA-Net and WHENet produce the highest yaw estimation errors. The Lightweight network once more is one of the most affected by occlusions, with the second highest average MAE. DAD-3D presents the lowest yaw estimation from SoTA methods, but one of the highest regarding pitch and roll. Our model also exhibit the lowest errors for this dataset with $\beta = 0.990$ presenting the lowest yaw error and $\beta = 0.999$ the lowest average MAE.  The error decreased in $1.04^\circ$ when compared to the best SoTA result (WHENet).

\subsection{Ablation study for several occlusion severities}
\begin{figure}[h]
 \centering
 \includegraphics[width=0.7\textwidth]{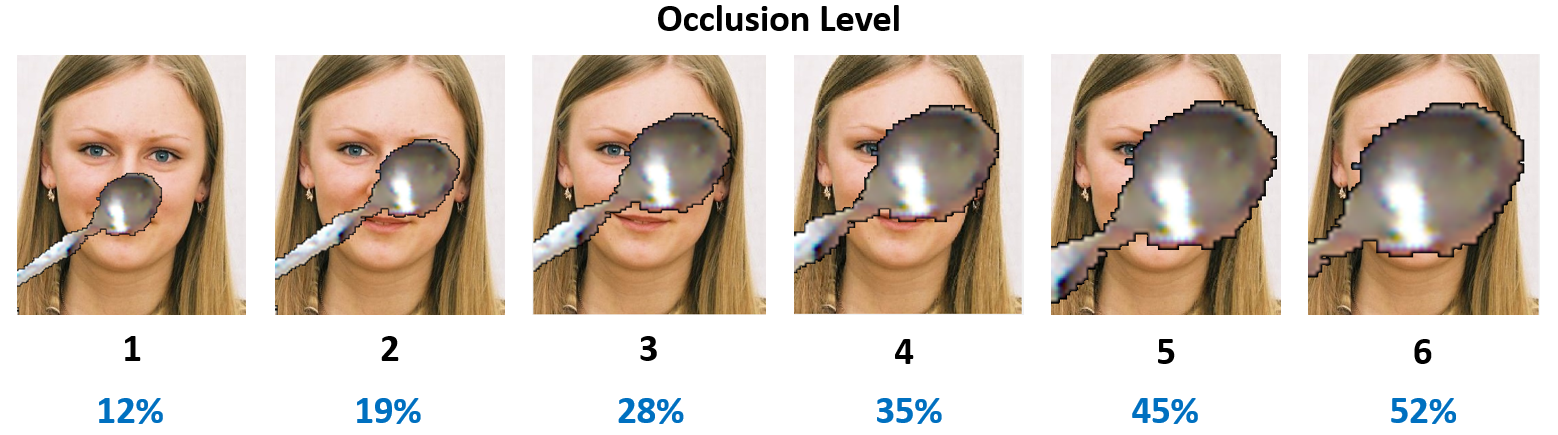}
 \caption{Occlusion severity levels.}
 \label{fig: occlusion_levels}
\end{figure}
\begin{figure}[h]
 \centering
 \includegraphics[width=0.9\textwidth]{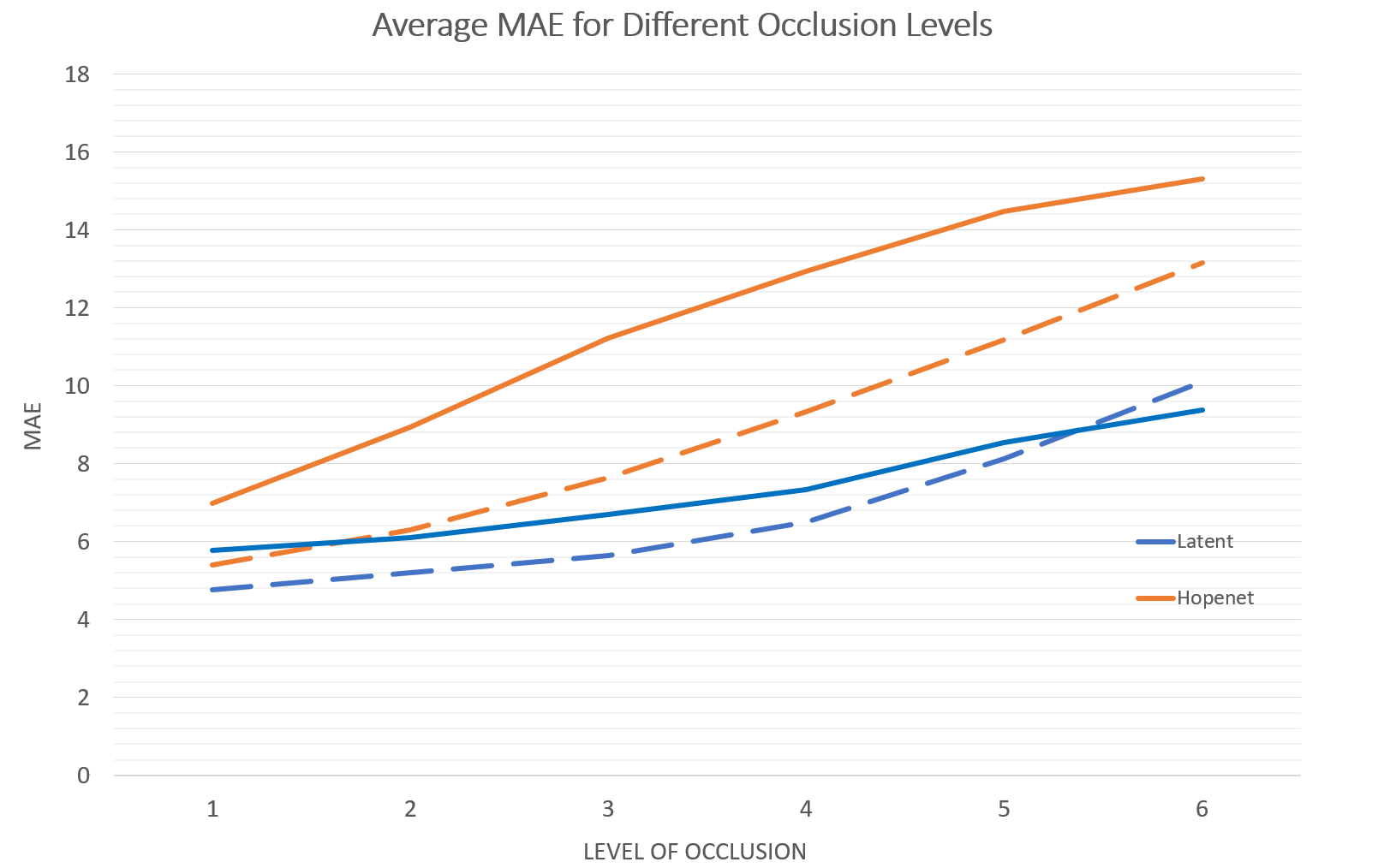}
 \caption{Average MAE comparison for occlusion dimension levels. Solid and dashed lines correspond to AFLW2000 and BIWI results, respectively.}
 \label{fig: biwi_levels}
\end{figure}

\begin{table}[h]
\begin{center}
\caption{Average MAE ($^\circ$) comparison for occlusion dimension levels.}\label{table:spoon results}
\resizebox{\columnwidth}{!}{
\begin{tabular}{c|c|c|c|c|c|c|c}
Datasets                  & Methods \textbackslash Occlusion level & 1     & 2     & 3      & 4      & 5      & 6      \\ \hline
\multirow{2}{*}{BIWI \cite{ref25}}     & Hopenet \cite{ref18}                                & 5.409 & 6.299 & 7.633  & 9.329  & 11.178 & 13.168 \\ \cline{2-8} 
                          & Latent (LSR)                           & 4.755 & 5.201 & 5.646  & 6.503  & 8.130  & 10.108 \\ \hline
\multirow{2}{*}{AFLW2000 \cite{ref24}} & Hopenet \cite{ref18}                                & 6.982 & 8.933 & 11.231 & 12.945 & 14.474 & 15.317 \\ \cline{2-8} 
                          & Latent (LSR)                           & 5.783 & 6.113 & 6.686  & 7.341  & 8.545  & 9.377 
\end{tabular}
}
\end{center}
\end{table}

In order to evaluate the robustness of our method against the occlusions, we performed an ablation study for different occlusion levels. To accomplish this, we simulated a feeding scenario. Concretely, we recorded an object, a spoon, in eleven different positions and orientations. This object is introduced in the images of both AFLW2000 and BIWI datasets, resulting in occluded versions of these two datasets. For this purpose, we applied our synthetic occlusion generation procedure (see Sec.~\ref{section: generating_occl}) and gradually augment the size of the spoon occlusions.
\par
We defined six levels of occlusions, ranging from one (lowest occlusion level) to six (highest occlusion level). As such, we generated six different occluded datasets for each AFLW2000 and BIWI datasets (twelve datasets are thus used in this experiment).
Examples from occlusion severity levels and respective average occlusion percentages for each dataset can be seen in Fig.~\ref{fig: occlusion_levels}. The percentages are computed as the proportion of pixels in the original dataset face images that become occluded.
We perform a comparison between the proposed methodology and Hopenet, as we next describe. The results depicted in Fig.~\ref{fig: biwi_levels} show that the difference in the estimation error between our model and Hopenet increases as we augment the occlusion (until level four). For the largest occlusions (level six) our model decreases the average estimation error in 39\% on AFLW2000 and 23\% on BIWI. Not only does it perform better with occlusions but it is also more robust to occlusion severity variations. The difference in the average MAE between levels one and six is around $8^\circ$ for Hopenet in both datasets, while for our model it is around $5^\circ$ on BIWI and lower than $4^\circ$ on AFLW2000. Furthermore, our model's error estimation for level six on AFLW2000 is lower than \textit{Hopenet's} error for level three occlusions and only slightly higher than level two. These results confirm that our model significantly improves the robustness of the estimation, even for very large occlusions.

\subsection{t-SNE latent space visualization}

\begin{figure}[h]
 \centering
 \includegraphics[width=1\textwidth]{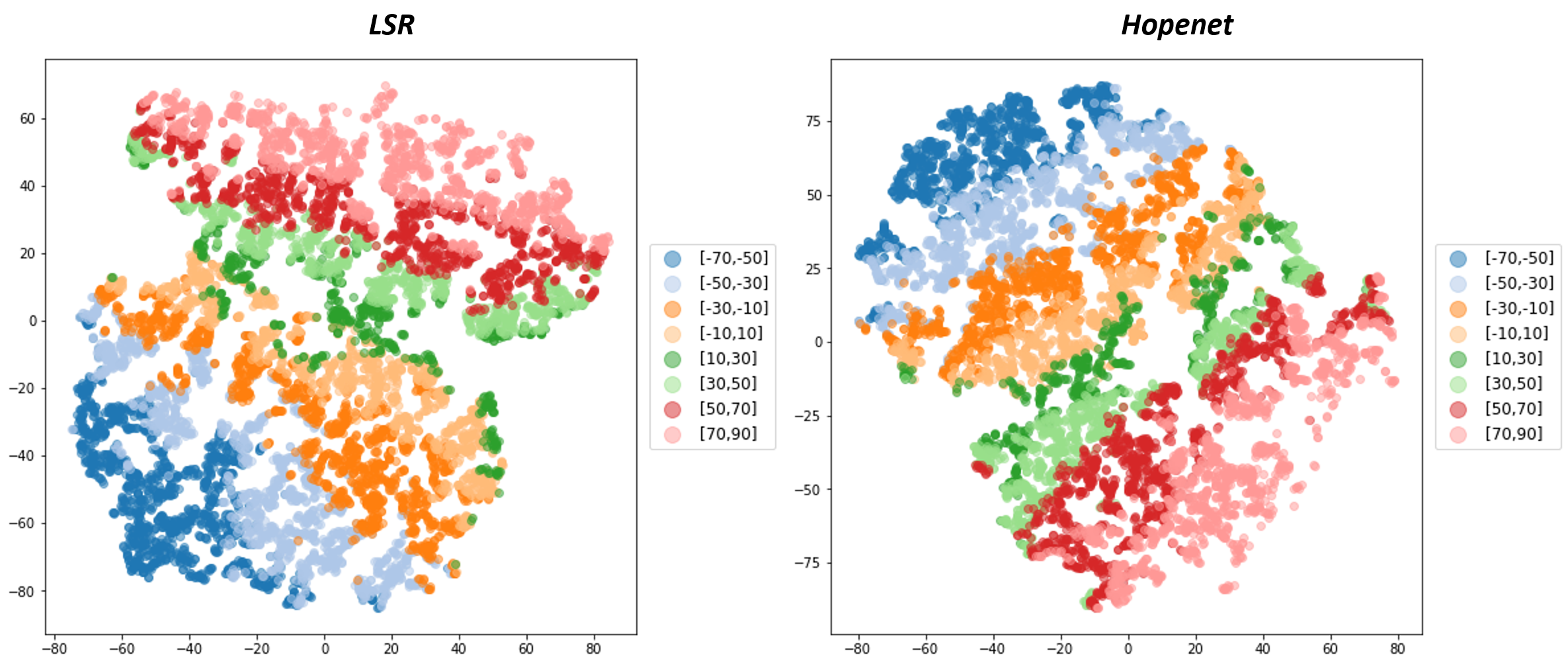}
 \caption{t-SNE latent space visualization - non-occluded images.  Data points from similar poses closer to each other and dissimilar poses far apart.  - good data arrangement. }
 \label{fig: tsne_non}
\end{figure}

\begin{figure}[h]
 \centering
 \includegraphics[width=1\textwidth]{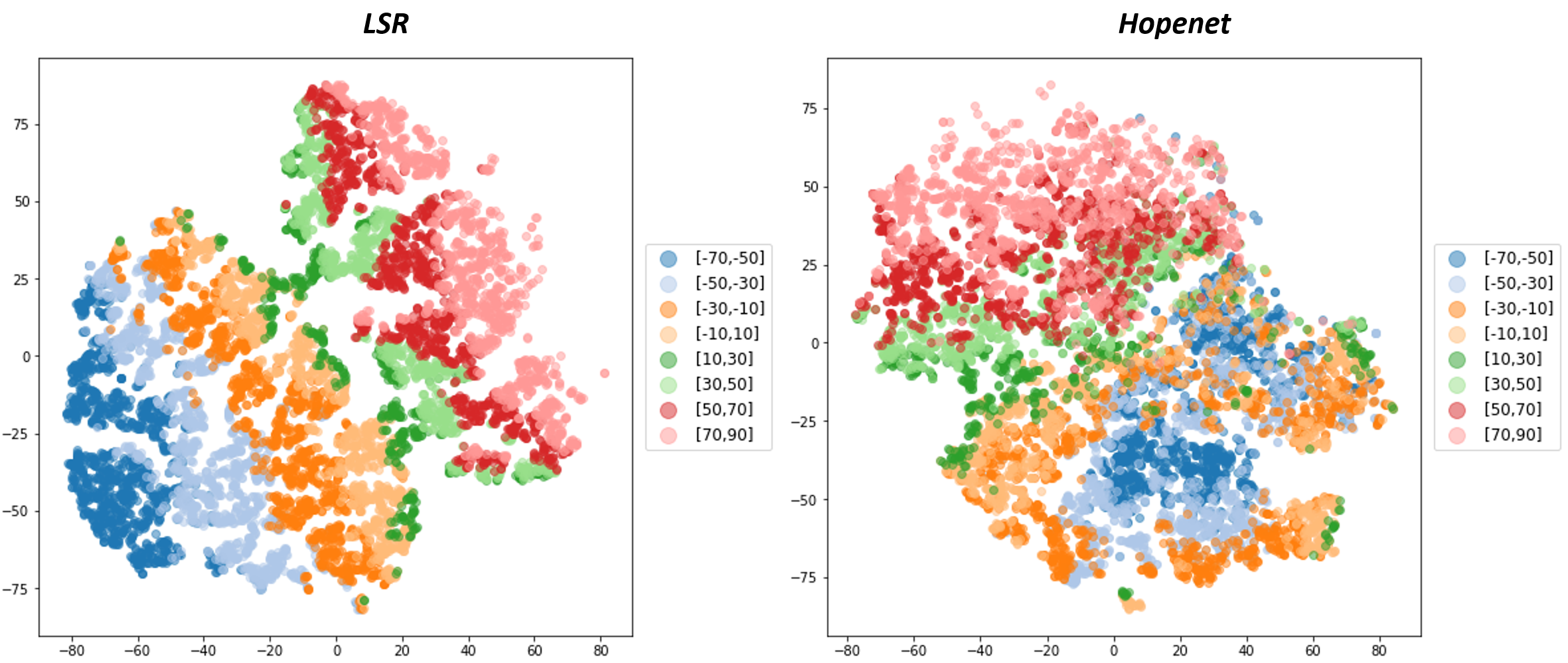}
 \caption{t-SNE latent space visualization - occluded images. Data points of different pose angles mixed for Hopenet. However, our model displays data points arranged in a continuous manner as it did for non-occluded images.}
 \label{fig: tsne_occ}
\end{figure}

 \par The latent space is an abstract multi-dimensional space that maps a relevant internal representation of how a neural network interprets external input data and provides insight on what learned from training. In this space, the distance of data points is related to the semantic similarity of the input data. As such, similar items are closer to one another in this space, while dissimilar items are further apart. 
We performed inference in non-occluded and occluded face images using both our model and Hopenet. The output latent embeddings were stored in order to apply the t-Distributed Stochastic Neighbor Embedding (t-SNE) technique. This is a non-linear method that allows to visualize high-dimensional data by giving each datapoint a location in a two or three-dimensional map.  In short, it is a visualization technique. The idea is to gain an intuition of how the models interpret and arrange both non-occluded and occluded data. In this space, a good HPE model should exhibit images of similar pose angles closer to each another and images of dissimilar poses further apart, with and without occlusions.
 \par The resultant t-SNE 2D maps can be seen in Figs.~\ref{fig: tsne_non},~\ref{fig: tsne_occ} for non-occluded and occluded images, respectively. 

\par  The t-SNE visualization for non-occluded images (see Fig.~\ref{fig: tsne_non}) displays a good data arrangement for both models, since the data points in the latent space are arranged from the lowest values of pose angles to the highest (and vice-versa) in a continuous progressive manner.  Similar pose angles are next to each other (e.g. angles in $[-70^\circ,-50^\circ[$ next to $[-50^\circ,-30^\circ[$, which in turn are next to $[-30^\circ,-10^\circ[$) and dissimilar are further apart (e.g. the most distant points from the $[-70^\circ,-50^\circ[$ range are in the $[70^\circ,90^\circ[$ range).  \par 
The t-SNE visualization for occluded images of Hopenet (see Fig.~\ref{fig: tsne_occ}) displays a much more erratic and ambiguous blend of image data points from different angles in several regions of the latent space. For instance, points in the lowest negative range ($[-70^\circ,-50^\circ[$) are now much closer to points in positive ranges (even to the most dissimilar and highest positive range, $[70^\circ,90^\circ[$) and very mixed with all negative ranges. In contrast, our model displays significantly better data arrangement for occlusions by presenting a continuous progression of angle values in the latent space, as it did for non-occluded images.
These t-SNE representations therefore provide an intuitive visual insight on how Hopenet struggles in distinguishing between different poses with occluded image inputs, while our model, aided by latent space regression, exhibits a good semantic understanding of head poses regardless of the face being occluded or not.

\subsection{A real case: the feeding robot scenario}

\begin{figure}[h]
 \centering
 \includegraphics[width=1\textwidth]{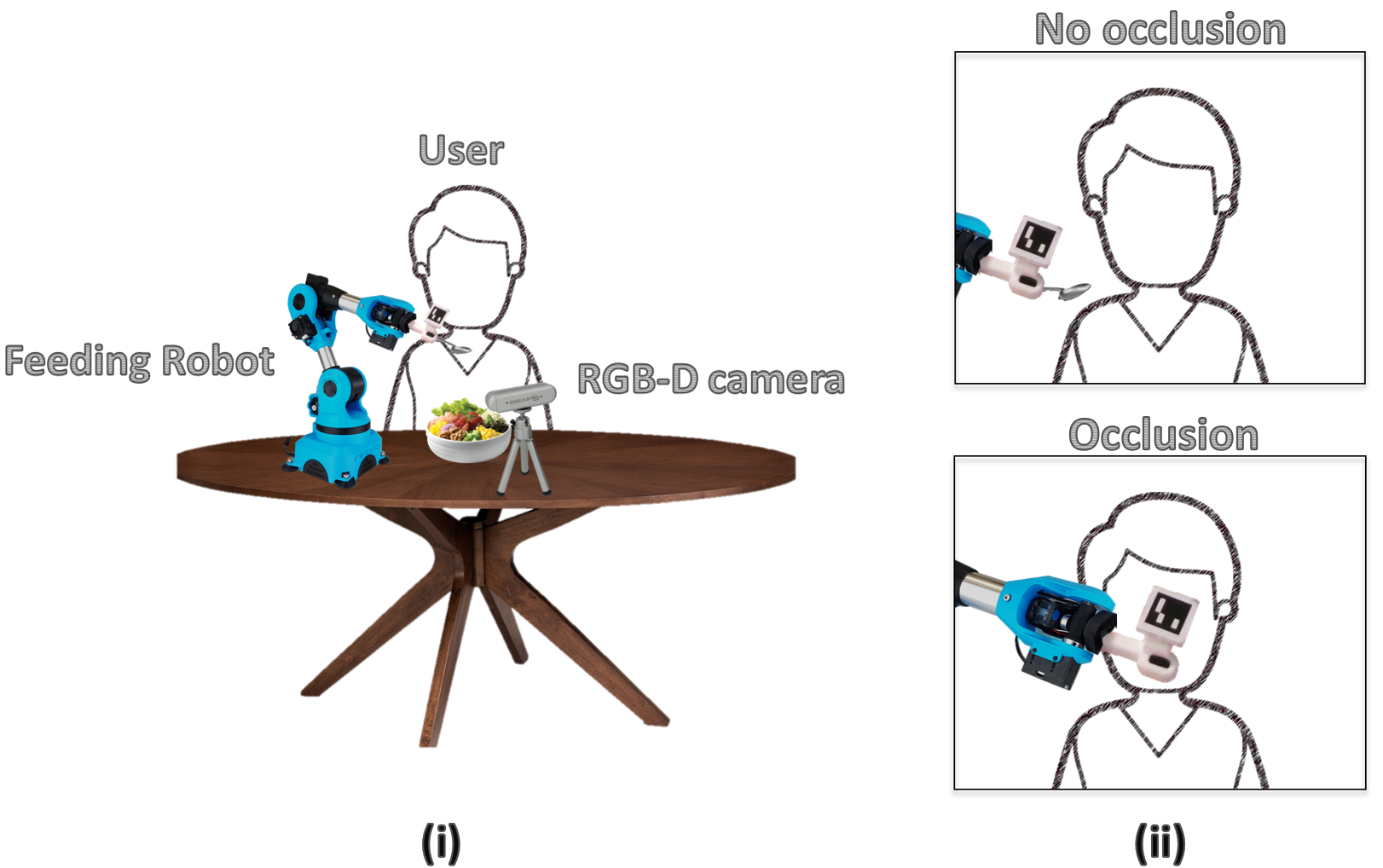}
 \caption{Feedbot's framework in autonomous feeding scenario (i) and camera point of view (ii).}
 \label{fig: feedbot_scenario}
\end{figure}


\begin{figure}[]
 \centering
 \includegraphics[width=1\textwidth]{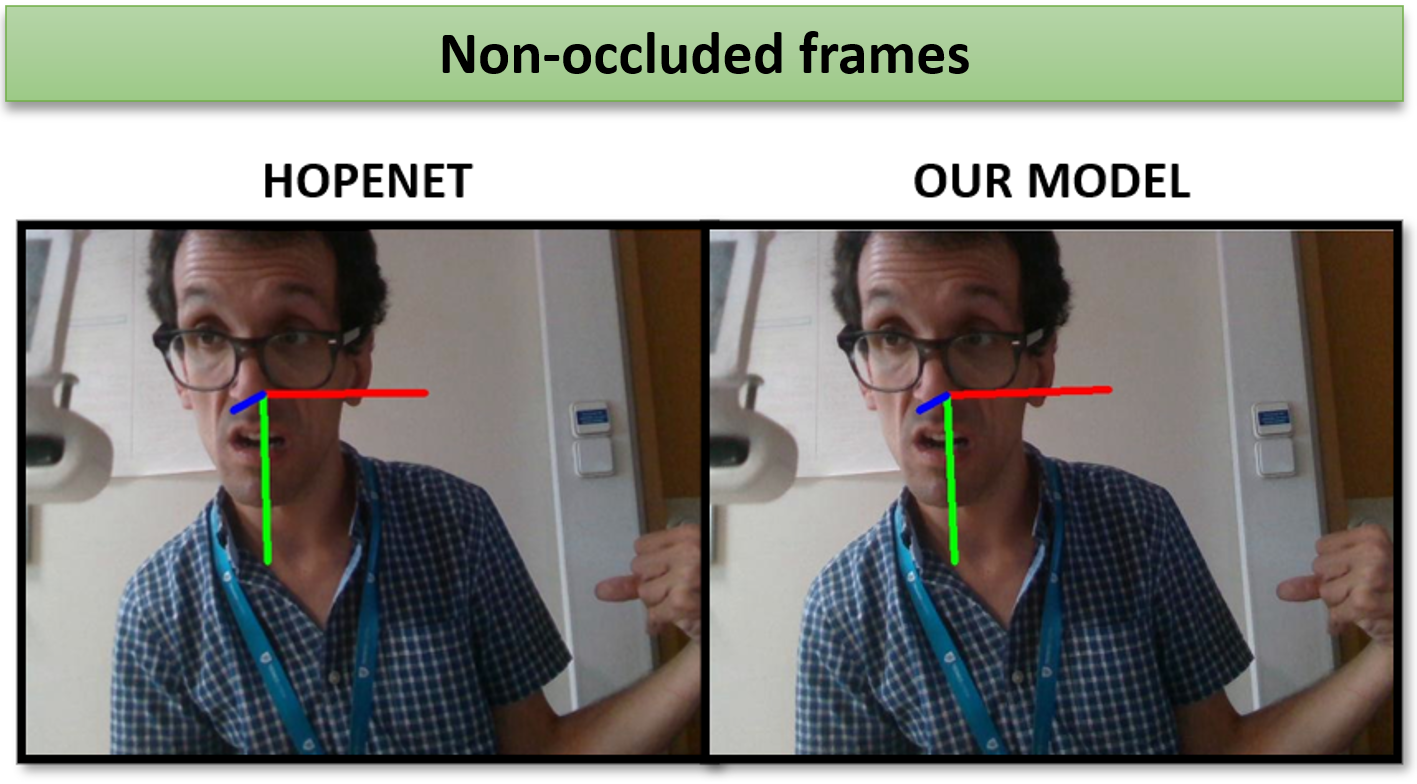}
 \caption{ Feedbot scenario: Comparison between Hopenet and our model. HPE is good and similar for both models in non-occluded frames.}
 \label{fig: feedbotcase}
\end{figure}

\begin{figure}[]
 \centering
 \includegraphics[width=1\textwidth]{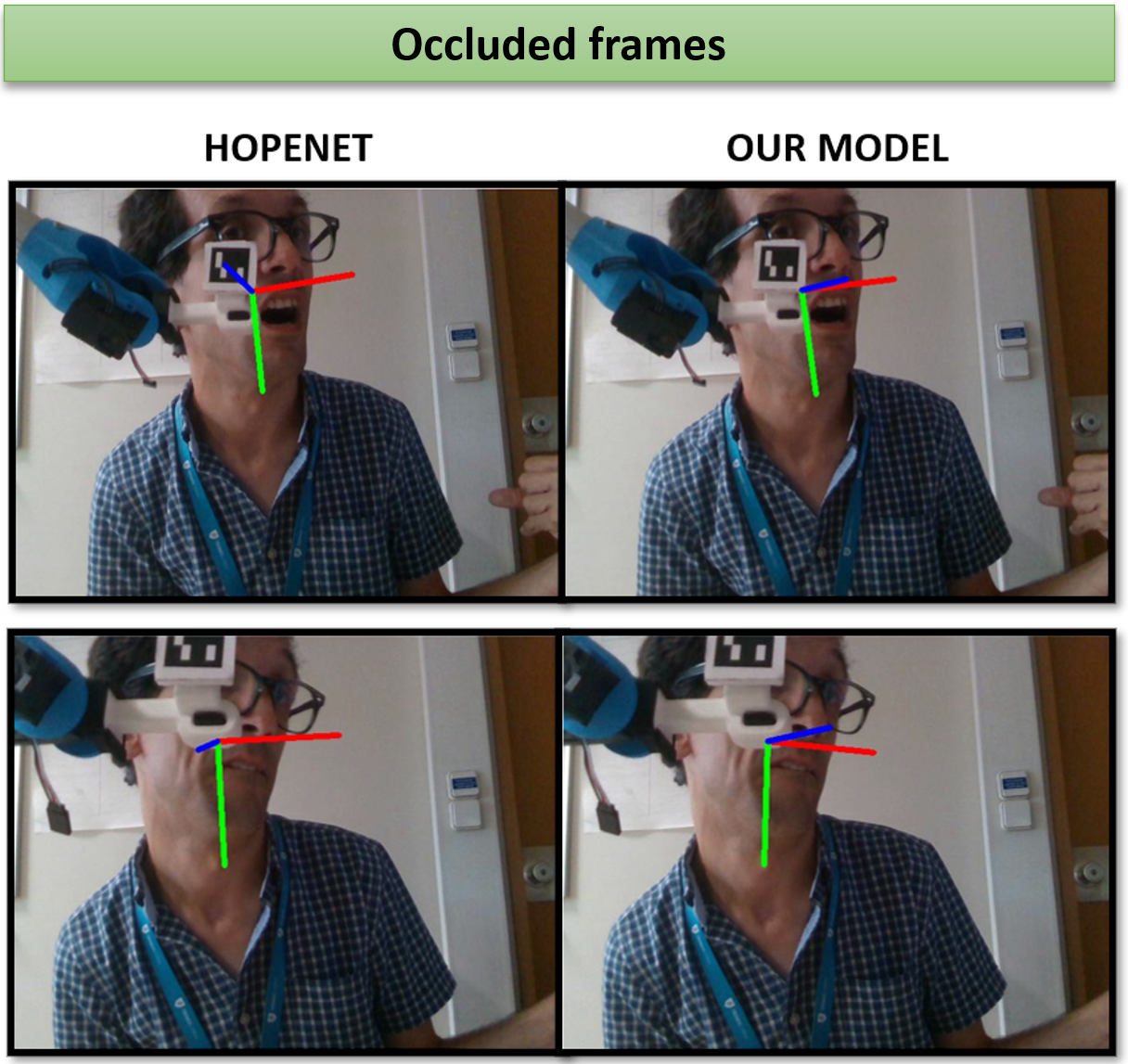}
 \caption{ Feedbot scenario: Comparison between Hopenet and our model. For occluded frames, however, Hopenet does not estimate the pose correctly. Our model correctly estimates the pose, despite the occlusions.}
 \label{fig: feedbotcase_o}
\end{figure}


We tested our best method in the feeding context of Feedbot ~\cite{ref41}, an autonomous feeding robot designed for people with upper-extremity disabilities. The Feedbot framework, illustrated in Fig.~\ref{fig: feedbot_scenario} (i), includes a RGB-D camera and a modular robotic arm to perform assisted feeding. The robotic arm is set close to the user to be able to perform the feeding task and the camera is placed in front of the user, but more distant. This way, the camera captures the entire environment, including the user and the robotic arm, as shown in Fig.~\ref{fig: feedbot_scenario} (ii). The end-effector of the robotic arm has a spoon to collect the food from the plate and a RGB camera that provides visual feedback on whether it was captured to improve food acquisition efficiency. A point cloud registration algorithm is implemented to use the 3D data captured by the RGB-D sensor and a 3D face model to compute rigid head transformations and track the head pose and mouth of the user. This tracking allows the food to be effectively brought to the user's mouth, rather than to a predefined feeding location. However, the robotic arm occludes the face while feeding, which makes the head pose estimation and tracking much more difficult. 
We recorded a video where Feedbot's robotic arm executes the feeding task and occludes the face of the user. We performed inference for the entire video using both our and \textit{Hopenet's} models, which exhibited the lowest error among the tested SoTA models in Section \ref{section 5.1.2}. Since we do not have the ground truth pose in this testing conditions, we carry out a qualitative analysis and evaluation in this section. 

\par Initially, when the robot is not yet feeding the user, the head is not occluded by the robotic arm and our model's pose estimation is accurate and identical to that of SoTA Hopenet, as seen in Fig.~\ref{fig: feedbotcase}. This is the desired behaviour since we intend to improve on occluded head pose estimation, while preserving the SoTA accuracy for non-occluded images.

\par When the robot moves the spoon towards the user's mouth, it occludes the face with the end-effector. Fig.~\ref{fig: feedbotcase_o} displays a set of occluded frames which show clear improvements in the head pose estimation of our model when compared to Hopenet. In these examples, Hopenet estimates head poses in opposite direction to that of which the head is turned. These results outline the difficulties that model has with the occlusions generated by the robot. Contrasting with the above, we verify that our model is noticeably more robust to this problem and estimates the head poses much closer to reality despite the partial occlusions, particularly improving the yaw rotation around the green axis (the most important pose rotation) when compared to Hopenet.

\section{Conclusion}
\label{sec: conclusion}

\par
In this work, we developed a learning-based methodology to deal with the occlusion problem in head pose estimation. To be able to implement and test this approach, we introduced an auxiliary procedure to generate synthetic occlusions in face images using an RGB-D camera. We show how to segment occlusions based on depth data captured by the camera and how to inpaint the occlusion in any RGB face image. We applied this procedure to three datasets and generated synthetically occluded versions for each dataset. 
\par We designed a new multi-loss head pose estimation framework combined with a latent space regression loss. We showed that introducing such latent space regression term in the loss is fundamental for the accuracy improvement and generalization for occluded images and non-occluded images. We performed an ablation study varying the occlusion severity and demonstrated that our model is far more robust to large occlusions when compared with Hopenet. We visualized the latent space representation using the t-SNE technique. We verified that the latent regression component in training allowed our model to achieve good data arrangement in regard to pose angles for non-occluded and occluded images particularly, which translated in better pose estimation. 
\par We carried out qualitative tests in the real world application of the Feedbot, an autonomous assisting feeding robot. Our model improved the head pose estimation for the occlusions of the robotic arm when compared to a SoTA estimation model, while achieving identical performance without occlusions. 

\subsection{Method limitations and future work}

Despite achieving good results, the developed methodologies have some limitations and further work could be done to improve them. 
\par The RGB camera of the Microsoft Kinect has low image resolution (640x480 pixels), which leads to synthetic occlusions of low definition. To improve the occluded datasets, an RGB-D sensor of higher resolution could be used to generate better defined synthetic occlusions.
\par The convolutional neural network we used in pose estimation frameworks, ResNet-50, is a large network with over 23 million parameters and is therefore slower to train and requires more GPU power. A lighter network such as EfficientNet \cite{ref10}, with 11 million parameters, could be used to address this issue and allow implementations in low-cost hardware. 
\par A method regarding the reconstruction of occluded faces could also be a good strategy to this problem. Particularly, a Generative Adversarial Network could be explored for the face generation. 










\bibliographystyle{elsarticle-num} 
\bibliography{reference.bib}

\begin{thebibliography}{10}
\expandafter\ifx\csname url\endcsname\relax
  \def\url#1{\texttt{#1}}\fi
\expandafter\ifx\csname urlprefix\endcsname\relax\def\urlprefix{URL }\fi
\expandafter\ifx\csname href\endcsname\relax
  \def\href#1#2{#2} \def\path#1{#1}\fi

\bibitem{ref1}
Y.~Zhou, J.~Gregson, Whenet: Real-time fine-grained estimation for wide range
  head pose (2020).
\newblock \href {http://arxiv.org/abs/2005.10353} {\path{arXiv:2005.10353}}.

\bibitem{ref44}
A.~Abate, C.~Bisogni, A.~Castiglione, M.~Nappi, Head pose estimation: An
  extensive survey on recent techniques and applications, Pattern Recognition
  127 (2022) 108591.
\newblock \href {https://doi.org/10.1016/j.patcog.2022.108591}
  {\path{doi:10.1016/j.patcog.2022.108591}}.

\bibitem{ref2}
A.~Fernández, R.~Usamentiaga, J.~L. Carús, R.~Casado, Driver distraction
  using visual-based sensors and algorithms, Sensors 16~(11) (2016).
\newblock \href {https://doi.org/10.3390/s16111805}
  {\path{doi:10.3390/s16111805}}.

\bibitem{ref38}
S.~J. Ray, J.~Teizer, Coarse head pose estimation of construction equipment
  operators to formulate dynamic blind spots, Adv. Eng. Inform. 26~(1) (2012)
  117–130.
\newblock \href {https://doi.org/10.1016/j.aei.2011.09.005}
  {\path{doi:10.1016/j.aei.2011.09.005}}.

\bibitem{ref3}
M.~C. d.~F. Macedo, A.~L. Apolinário, A.~C. d.~S. Souza, A robust real-time
  face tracking using head pose estimation for a markerless ar system, in: 2013
  XV Symposium on Virtual and Augmented Reality, 2013, pp. 224--227.
\newblock \href {https://doi.org/10.1109/SVR.2013.12}
  {\path{doi:10.1109/SVR.2013.12}}.

\bibitem{ref35}
F.~Nihei, Y.~I. Nakano, Exploring methods for predicting important utterances
  contributing to meeting summarization, Multimodal Technologies and
  Interaction 3 (2019) 50.
\newblock \href {https://doi.org/10.3390/mti3030050}
  {\path{doi:10.3390/mti3030050}}.

\bibitem{ref36}
I.~Reid, B.~Benfold, A.~Patron-Perez, E.~Sommerlade, Understanding interactions
  and guiding visual surveillance by tracking attention, Vol. 6468, 2010, pp.
  380--389.
\newblock \href {https://doi.org/10.1007/978-3-642-22822-3\textunderscore38}
  {\path{doi:10.1007/978-3-642-22822-3\textunderscore38}}.

\bibitem{ref37}
C.-W. Chen, H.~Aghajan, Multiview social behavior analysis in work
  environments, 2011, pp. 1 -- 6.
\newblock \href {https://doi.org/10.1109/ICDSC.2011.6042910}
  {\path{doi:10.1109/ICDSC.2011.6042910}}.

\bibitem{ref39}
K.~Sankaranarayanan, M.-C. Chang, N.~Krahnstoever, Tracking gaze direction from
  far-field surveillance cameras, in: 2011 IEEE Workshop on Applications of
  Computer Vision (WACV), 2011, pp. 519--526.
\newblock \href {https://doi.org/10.1109/WACV.2011.5711548}
  {\path{doi:10.1109/WACV.2011.5711548}}.

\bibitem{ref40}
K.~Smith, S.~O. Ba, J.-M. Odobez, D.~Gatica-Perez, Tracking the visual focus of
  attention for a varying number of wandering people, IEEE Transactions on
  Pattern Analysis and Machine Intelligence 30~(7) (2008) 1212--1229.
\newblock \href {https://doi.org/10.1109/TPAMI.2007.70773}
  {\path{doi:10.1109/TPAMI.2007.70773}}.

\bibitem{ref49}
T.~Hassner, S.~Harel, E.~Paz, R.~Enbar, Effective face frontalization in
  unconstrained images (11 2014).
\newblock \href {https://doi.org/10.1109/CVPR.2015.7299058}
  {\path{doi:10.1109/CVPR.2015.7299058}}.

\bibitem{ref6}
J.~M. Diaz~Barros, B.~Mirbach, F.~Garcia, K.~Varanasi, D.~Stricker, Real-Time
  Head Pose Estimation by Tracking and Detection of Keypoints and Facial
  Landmarks, 2019, pp. 326--349.
\newblock \href {https://doi.org/10.1007/978-3-030-26756-8\textunderscore16}
  {\path{doi:10.1007/978-3-030-26756-8\textunderscore16}}.

\bibitem{ref65}
H.~Yuan, M.~Li, J.~Hou, J.~Xiao, Single image-based head pose estimation with
  spherical parametrization and 3d morphing, Pattern Recognition 103 (2020)
  107316.
\newblock \href {https://doi.org/10.1016/j.patcog.2020.107316}
  {\path{doi:10.1016/j.patcog.2020.107316}}.

\bibitem{dad3d}
T.~Martyniuk, O.~Kupyn, Y.~Kurlyak, I.~Krashenyi, J.~Matas, V.~Sharmanska,
  Dad-3dheads: A large-scale dense, accurate and diverse dataset for 3d head
  alignment from a single image, in: Proc. IEEE Conf. on Computer Vision and
  Pattern Recognition (CVPR), 2022.

\bibitem{ref11}
J.~Guo, X.~Zhu, Y.~Yang, F.~Yang, Z.~Lei, S.~Z. Li, Towards fast, accurate and
  stable 3d dense face alignment, in: A.~Vedaldi, H.~Bischof, T.~Brox, J.-M.
  Frahm (Eds.), Computer Vision -- ECCV 2020, Springer International
  Publishing, Cham, 2020, pp. 152--168.

\bibitem{ref14}
M.~Wenzel, W.~Schiffmann, Head pose estimation of partially occluded faces,
  2005, pp. 353-- 360.
\newblock \href {https://doi.org/10.1109/CRV.2005.45}
  {\path{doi:10.1109/CRV.2005.45}}.

\bibitem{ref16}
Y.~Wu, Q.~Ji, Robust facial landmark detection under significant head poses and
  occlusion, 2015 IEEE International Conference on Computer Vision (ICCV)
  (2015) 3658--3666.

\bibitem{ref17}
Y.~Wu, C.~Gou, Q.~Ji, Simultaneous facial landmark detection, pose and
  deformation estimation under facial occlusion, 2017 IEEE Conference on
  Computer Vision and Pattern Recognition (CVPR) (2017) 5719--5728.

\bibitem{ref18}
N.~Ruiz, E.~Chong, J.~M. Rehg, Fine-grained head pose estimation without
  keypoints, 2018 IEEE/CVF Conference on Computer Vision and Pattern
  Recognition Workshops (CVPRW) (2018) 2155--215509.

\bibitem{ref64}
Y.~Xu, C.~Jung, Y.~Chang, Head pose estimation using deep neural networks and
  3d point cloud, Pattern Recognition 121 (2021) 108210.
\newblock \href {https://doi.org/10.1016/j.patcog.2021.108210}
  {\path{doi:10.1016/j.patcog.2021.108210}}.

\bibitem{ref19}
T.-Y. Yang, Y.-T. Chen, Y.-Y. Lin, Y.-Y. Chuang, Fsa-net: Learning fine-grained
  structure aggregation for head pose estimation from a single image, in: 2019
  IEEE/CVF Conference on Computer Vision and Pattern Recognition (CVPR), 2019,
  pp. 1087--1096.
\newblock \href {https://doi.org/10.1109/CVPR.2019.00118}
  {\path{doi:10.1109/CVPR.2019.00118}}.

\bibitem{ref20}
T.-Y. Yang, Y.-H. Huang, Y.-Y. Lin, P.-C. Hsiu, Y.-Y. Chuang, Ssr-net: A
  compact soft stagewise regression network for age estimation, in: Proceedings
  of the Twenty-Seventh International Joint Conference on Artificial
  Intelligence, {IJCAI-18}, International Joint Conferences on Artificial
  Intelligence Organization, 2018, pp. 1078--1084.
\newblock \href {https://doi.org/10.24963/ijcai.2018/150}
  {\path{doi:10.24963/ijcai.2018/150}}.

\bibitem{ref21}
V.~Albiero, X.~Chen, X.~Yin, G.~Pang, T.~Hassner, img2pose: Face alignment and
  detection via 6dof, face pose estimation, 2021 IEEE/CVF Conference on
  Computer Vision and Pattern Recognition (CVPR) (2021) 7613--7623.

\bibitem{quatnet}
H.-W. Hsu, T.-Y. Wu, S.~Wan, W.~H. Wong, C.-Y. Lee, Quatnet: Quaternion-based
  head pose estimation with multiregression loss, IEEE Transactions on
  Multimedia 21~(4) (2019) 1035--1046.
\newblock \href {https://doi.org/10.1109/TMM.2018.2866770}
  {\path{doi:10.1109/TMM.2018.2866770}}.

\bibitem{6drepnet}
T.~Hempel, A.~A. Abdelrahman, A.~Al-Hamadi, 6d rotation representation for
  unconstrained head pose estimation, in: 2022 IEEE International Conference on
  Image Processing (ICIP), 2022, pp. 2496--2500.
\newblock \href {https://doi.org/10.1109/ICIP46576.2022.9897219}
  {\path{doi:10.1109/ICIP46576.2022.9897219}}.

\bibitem{lightweight}
X.~Li, D.~Zhang, M.~Li, D.-J. Lee, Accurate head pose estimation using image
  rectification and a lightweight convolutional neural network, IEEE
  Transactions on Multimedia (2022) 1--1\href
  {https://doi.org/10.1109/TMM.2022.3144893}
  {\path{doi:10.1109/TMM.2022.3144893}}.

\bibitem{ref51}
H.-W. Hsu, T.-Y. Wu, S.~Wan, W.~H. Wong, C.-Y. Lee, Quatnet: Quaternion-based
  head pose estimation with multiregression loss, IEEE Transactions on
  Multimedia 21~(4) (2019) 1035--1046.
\newblock \href {https://doi.org/10.1109/TMM.2018.2866770}
  {\path{doi:10.1109/TMM.2018.2866770}}.

\bibitem{ref53}
Y.~Wang, G.~Yuan, X.~Fu, Driver’s head pose and gaze zone estimation based on
  multi-zone templates registration and multi-frame point cloud fusion, Sensors
  22 (2022) 3154.
\newblock \href {https://doi.org/10.3390/s22093154}
  {\path{doi:10.3390/s22093154}}.

\bibitem{ref56}
T.~Baltru{\v s}aitis, P.~Robinson, L.-P. Morency, 3d constrained local model
  for rigid and non-rigid facial tracking, 2012 IEEE Conference on Computer
  Vision and Pattern Recognition (2012) 2610--2617.

\bibitem{ref25}
K.~S. Mader, {Biwi Kinect Head Pose Database} (2018).

\bibitem{ref57}
M.~Ariz, J.~J. Bengoechea, A.~Villanueva, R.~Cabeza, A novel 2d/3d database
  with automatic face annotation for head tracking and pose estimation.,
  Comput. Vis. Image Underst. 148 (2016) 201--210.

\bibitem{ref58}
M.~La~Cascia, S.~Sclaroff, V.~Athitsos, Fast, reliable head tracking under
  varying illumination: an approach based on registration of texture-mapped 3d
  models, IEEE Transactions on Pattern Analysis and Machine Intelligence 22~(4)
  (2000) 322--336.
\newblock \href {https://doi.org/10.1109/34.845375}
  {\path{doi:10.1109/34.845375}}.

\bibitem{ref24}
X.~Zhu, X.~Liu, Z.~Lei, S.~Z. Li, Face alignment in full pose range: A 3d total
  solution, IEEE Transactions on Pattern Analysis and Machine Intelligence
  41~(1) (2019) 78–92.
\newblock \href {https://doi.org/10.1109/tpami.2017.2778152}
  {\path{doi:10.1109/tpami.2017.2778152}}.

\bibitem{ref26}
X.~Yin, X.~Yu, K.~Sohn, X.~Liu, M.~Chandraker, Towards large-pose face
  frontalization in the wild, 2017 IEEE International Conference on Computer
  Vision (ICCV) (2017) 4010--4019.

\bibitem{ref59}
S.~Colaco, D.~S. Han, Facial keypoint detection with convolutional neural
  networks, in: 2020 International Conference on Artificial Intelligence in
  Information and Communication (ICAIIC), 2020, pp. 671--674.
\newblock \href {https://doi.org/10.1109/ICAIIC48513.2020.9065279}
  {\path{doi:10.1109/ICAIIC48513.2020.9065279}}.

\bibitem{ref60}
H.~Proença, J.~Neves, T.~Marques, S.~Barra, J.~Moreno, Joint head pose / soft
  label estimation for human recognition in-the-wild, IEEE Transactions on
  Pattern Analysis and Machine Intelligence 38 (2016) 1--1.
\newblock \href {https://doi.org/10.1109/TPAMI.2016.2522441}
  {\path{doi:10.1109/TPAMI.2016.2522441}}.

\bibitem{ref61}
J.~Deng, J.~Guo, N.~Xue, S.~Zafeiriou, Arcface: Additive angular margin loss
  for deep face recognition, in: 2019 IEEE/CVF Conference on Computer Vision
  and Pattern Recognition (CVPR), 2019, pp. 4685--4694.
\newblock \href {https://doi.org/10.1109/CVPR.2019.00482}
  {\path{doi:10.1109/CVPR.2019.00482}}.

\bibitem{ref66}
K.~Chen, K.~Jia, H.~Huttunen, J.~Matas, J.-K. Kämäräinen, Cumulative
  attribute space regression for head pose estimation and color constancy,
  Pattern Recognition 87 (10 2018).
\newblock \href {https://doi.org/10.1016/j.patcog.2018.10.015}
  {\path{doi:10.1016/j.patcog.2018.10.015}}.

\bibitem{ref62}
P.~Barra, C.~Bisogni, M.~Nappi, S.~Ricciardi, Fast QuadTree-Based Pose
  Estimation for Security Applications Using Face Biometrics: 12th
  International Conference, NSS 2018, Hong Kong, China, August 27-29, 2018,
  Proceedings, 2018, pp. 160--173.
\newblock \href {https://doi.org/10.1007/978-3-030-02744-5\textunderscore12}
  {\path{doi:10.1007/978-3-030-02744-5\textunderscore12}}.

\bibitem{ref63}
A.~F. Abate, P.~Barra, C.~Bisogni, M.~Nappi, S.~Ricciardi, Near real-time three
  axis head pose estimation without training, IEEE Access 7 (2019)
  64256--64265.

\bibitem{ref55}
G.~Borghi, M.~Venturelli, R.~Vezzani, R.~Cucchiara, Poseidon: Face-from-depth
  for driver pose estimation, in: 2017 IEEE Conference on Computer Vision and
  Pattern Recognition (CVPR), IEEE, 2017, pp. 5494--5503.

\bibitem{ref46}
I.~Khanfir, S.~Almouahed, B.~Solaiman, E.~Boss\'e, An iterative possibilistic
  knowledge diffusion approach for blind medical image segmentation, Pattern
  Recognition 78 (06 2018).
\newblock \href {https://doi.org/10.1016/j.patcog.2018.01.024}
  {\path{doi:10.1016/j.patcog.2018.01.024}}.

\bibitem{ref47}
G.~Ros, L.~Sellart, J.~Materzynska, D.~Vazquez, A.~M. Lopez, The synthia
  dataset: A large collection of synthetic images for semantic segmentation of
  urban scenes, in: 2016 IEEE Conference on Computer Vision and Pattern
  Recognition (CVPR), 2016, pp. 3234--3243.
\newblock \href {https://doi.org/10.1109/CVPR.2016.352}
  {\path{doi:10.1109/CVPR.2016.352}}.

\bibitem{ref45}
Y.~Wang, W.~Liang, J.~Shen, Y.~Jia, L.-F. Yu, A deep coarse-to-fine network for
  head pose estimation from synthetic data, Pattern Recognition 94 (05 2019).
\newblock \href {https://doi.org/10.1016/j.patcog.2019.05.026}
  {\path{doi:10.1016/j.patcog.2019.05.026}}.

\bibitem{ref41}
A.~Candeias, T.~Rhodes, M.~Marques, J.~a.~P. ao~Costeira, M.~Veloso, Vision
  augmented robot feeding, in: Proceedings of the European Conference on
  Computer Vision (ECCV) Workshops, 2018.

\bibitem{ref10}
M.~Tan, Q.~Le, {E}fficient{N}et: Rethinking model scaling for convolutional
  neural networks, in: K.~Chaudhuri, R.~Salakhutdinov (Eds.), Proceedings of
  the 36th International Conference on Machine Learning, Vol.~97 of Proceedings
  of Machine Learning Research, PMLR, 2019, pp. 6105--6114,
  \url{https://proceedings.mlr.press/v97/tan19a.html}.

\end{thebibliography}


\end{document}